\documentclass[10pt,twocolumn,letterpaper]{article}

\usepackage{iccv}
\usepackage{times}
\usepackage{epsfig}
\usepackage{graphicx}
\usepackage{amsmath}
\usepackage{amssymb}
\usepackage{multirow}
\usepackage{tkz-euclide}
\usepackage{subcaption}
\usepackage{comment}
\usepackage{booktabs}
\usepackage{tabularx}
\usepackage{makecell}
\definecolor{bbox}{RGB}{48, 188, 227}
\definecolor{rbbox}{RGB}{121, 224, 167}
\definecolor{bfov}{RGB}{224, 161, 101}
\definecolor{rbfov}{RGB}{224, 107, 96}
\definecolor{sec}{RGB}{39, 87, 146}
\definecolor{fir}{RGB}{199, 85, 93}
\usepackage{gensymb}
\usepackage{enumitem}
\usepackage{arydshln}
\usepackage{algorithm}
\usepackage{algorithmicx}
\usepackage[noend]{algpseudocode}
\usepackage[toc,page]{appendix}
\usepackage{tabularx}
\newcolumntype{H}{>{\setbox0=\hbox\bgroup}c<{\egroup}@{}}

\makeatletter
\newcommand{\multiline}[1]{%
  \begin{tabularx}{\dimexpr\linewidth-\ALG@thistlm}[t]{@{}X@{}}
    #1
  \end{tabularx}
}
\makeatother

\newcommand{\st}[1]{\textcolor{fir}{\bf #1}} 
\newcommand{\nd}[1]{\textcolor{sec}{\bf #1}} 
\newcommand{\edited}[1]{\textcolor{black}{#1}}

\usepackage[pagebackref=true,breaklinks=true,letterpaper=true,colorlinks,bookmarks=false]{hyperref}

\iccvfinalcopy 

\def\iccvPaperID{9582} 

\ificcvfinal\fi

\title{360VOT: A New Benchmark Dataset for Omnidirectional Visual Object Tracking}
\author{Huajian Huang
\and
Yinzhe Xu
\and
Yingshu Chen
\and
Sai-Kit Yeung
\and
The Hong Kong University of Science and Technology\\
{\tt\small \{hhuangbg, yxuck, ychengw\}@connect.ust.hk, saikit@ust.hk}\\
}

\begin{document}
\twocolumn[{%
\renewcommand\twocolumn[1][]{#1}%
\maketitle
\begin{center}
    \vspace{-1.9em} 
    \centering
    \captionsetup{type=figure}
    \def\imgw{0.23}
    \def\imgh{0.11}
    
    \subfloat{\includegraphics[width=\imgw\linewidth, height=\imgh\linewidth]{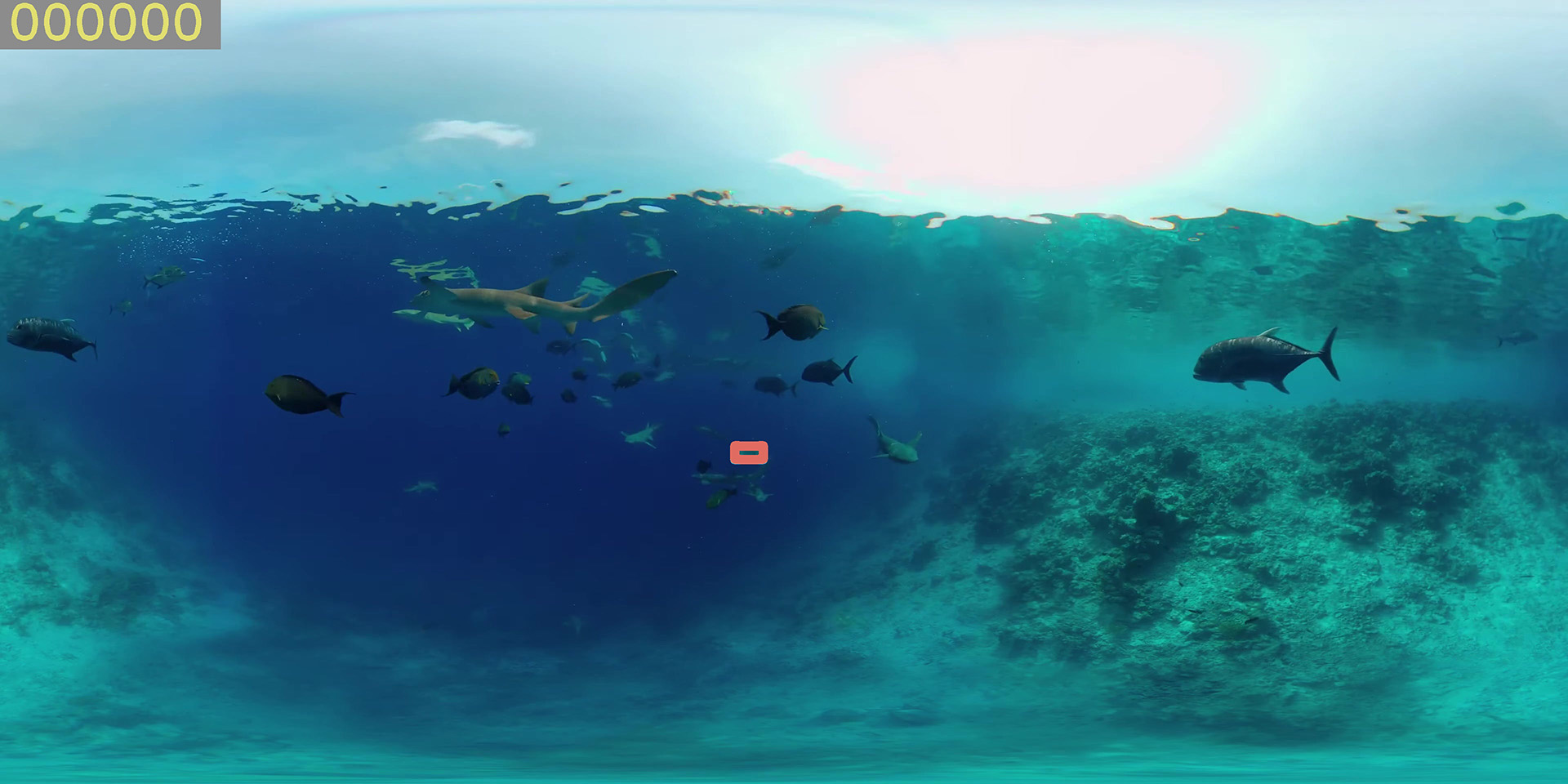}}\!
    \subfloat{\includegraphics[width=\imgw\linewidth, height=\imgh\linewidth]{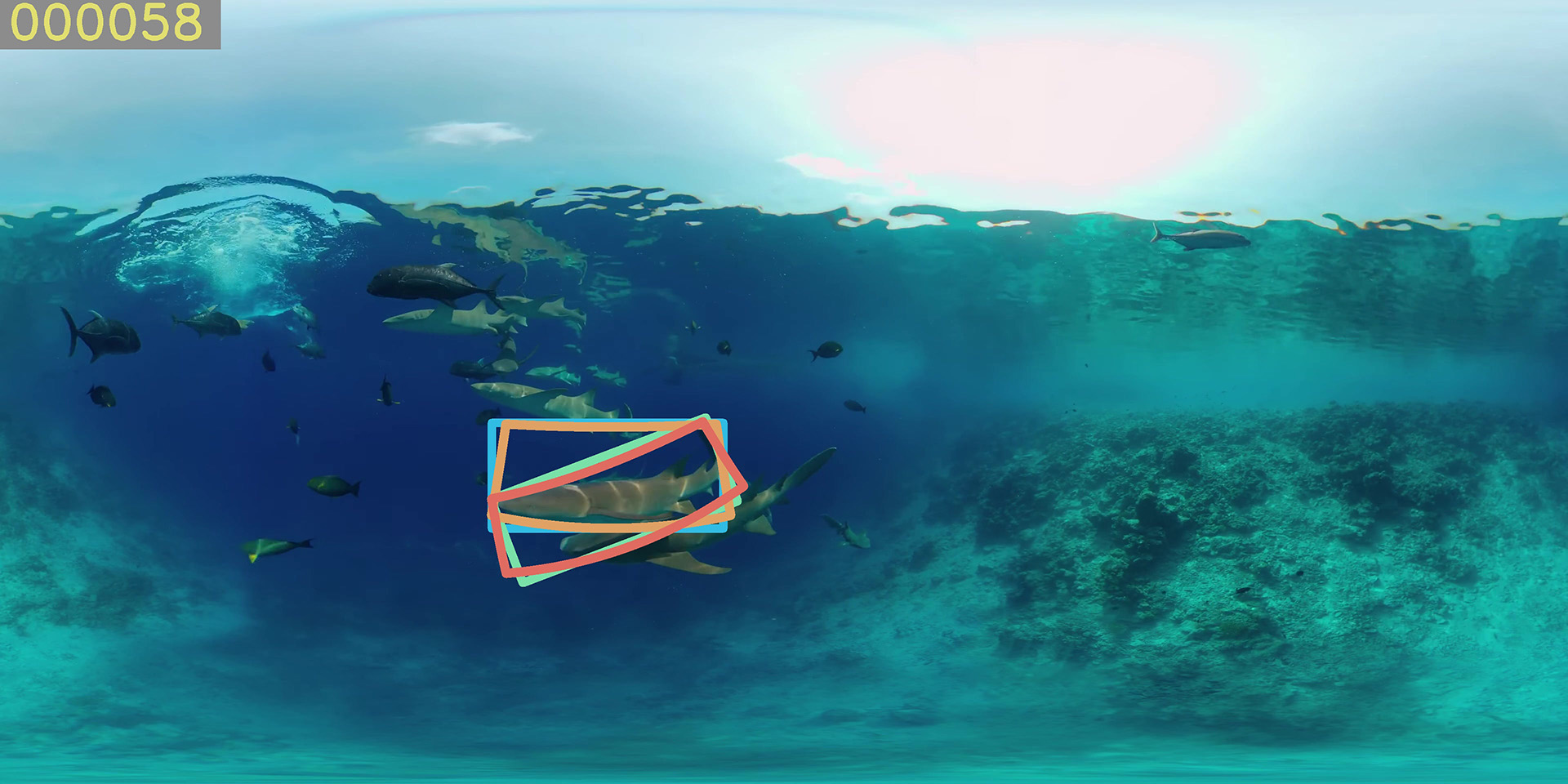}}\!
    \subfloat{\includegraphics[width=\imgw\linewidth, height=\imgh\linewidth]{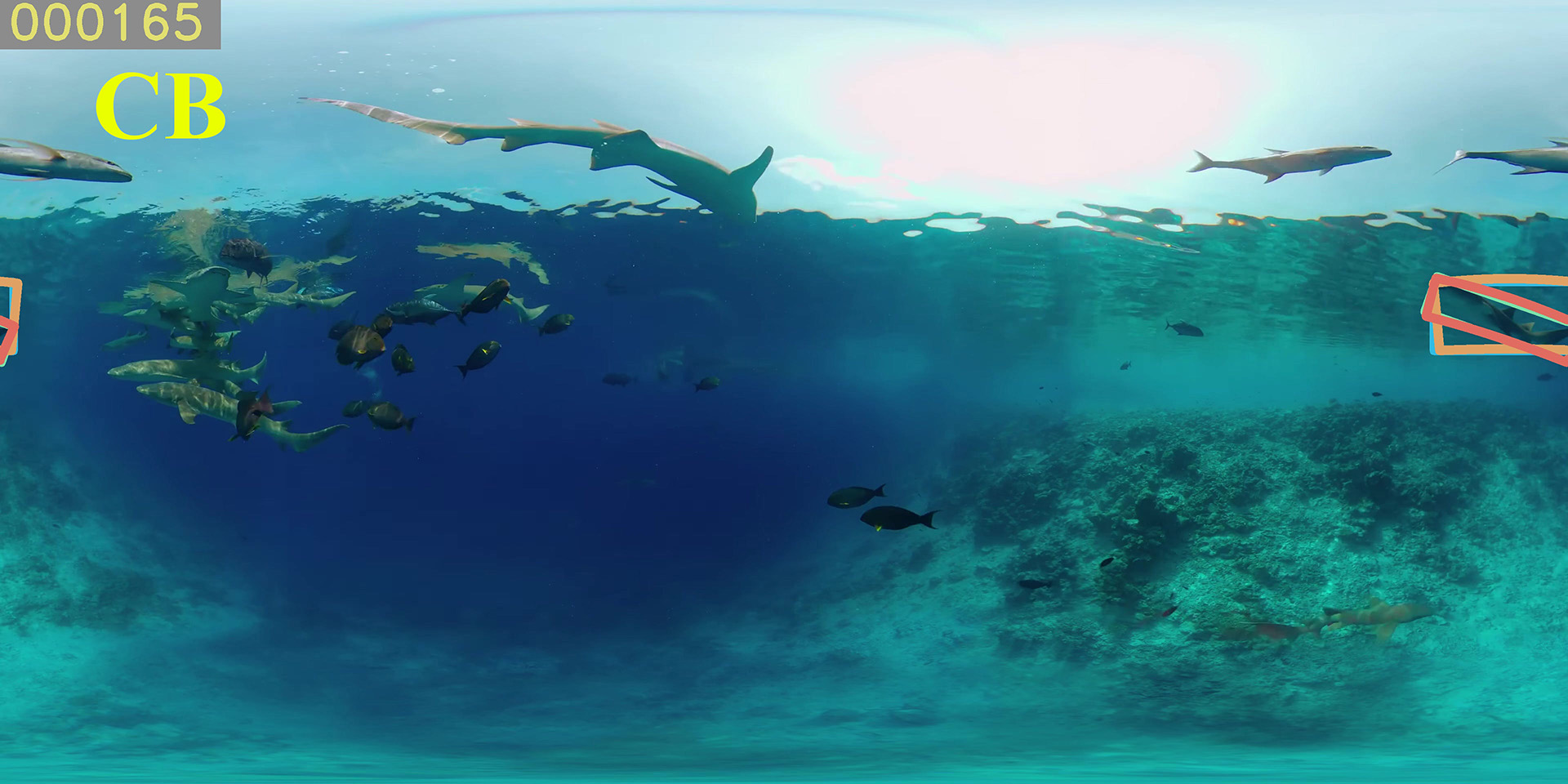}}\!
    \subfloat{\includegraphics[width=\imgw\linewidth, height=\imgh\linewidth]{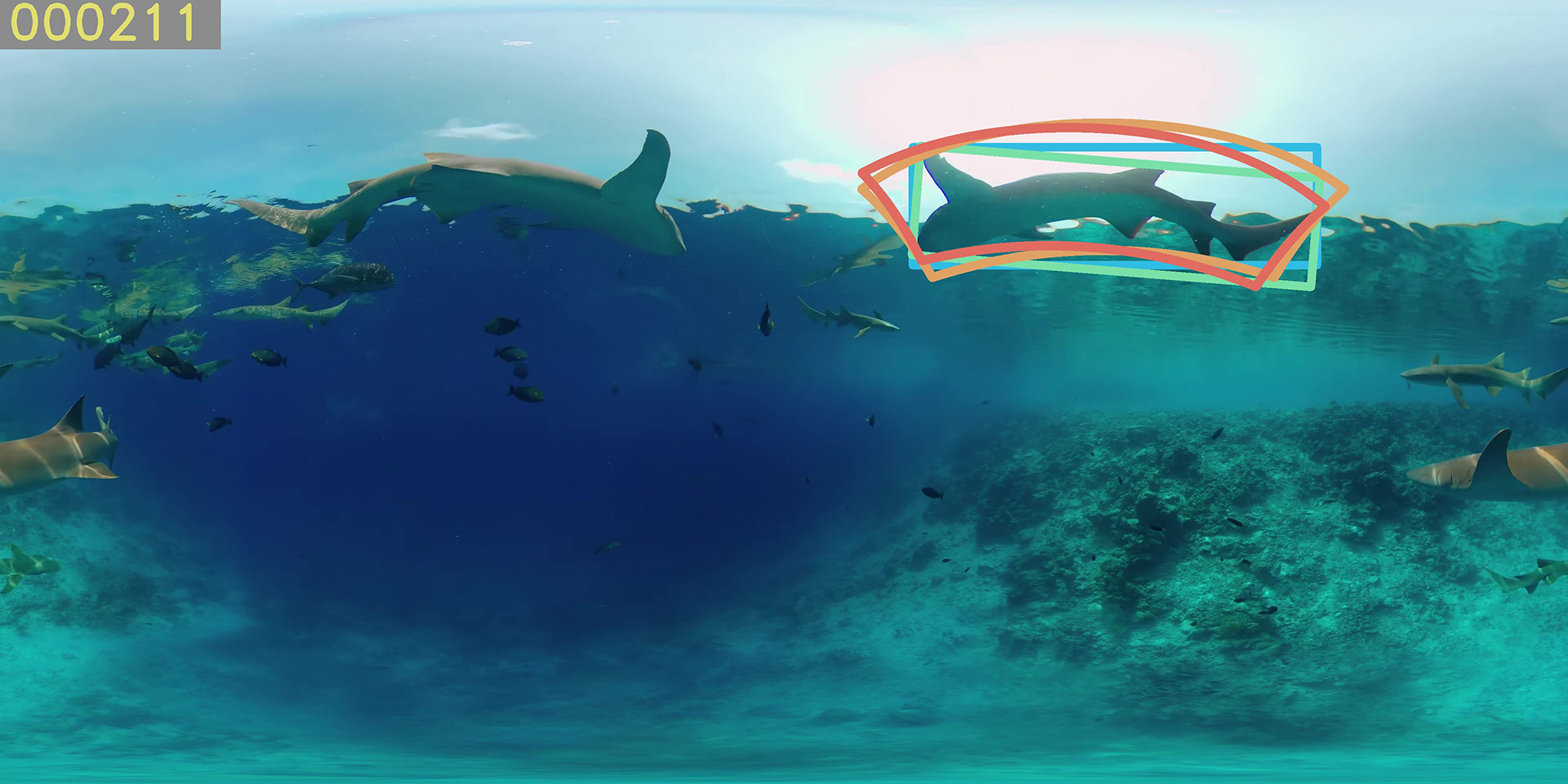}}
    \vspace{0.05em}
    
    \subfloat{\includegraphics[width=\imgw\linewidth, height=\imgh\linewidth]{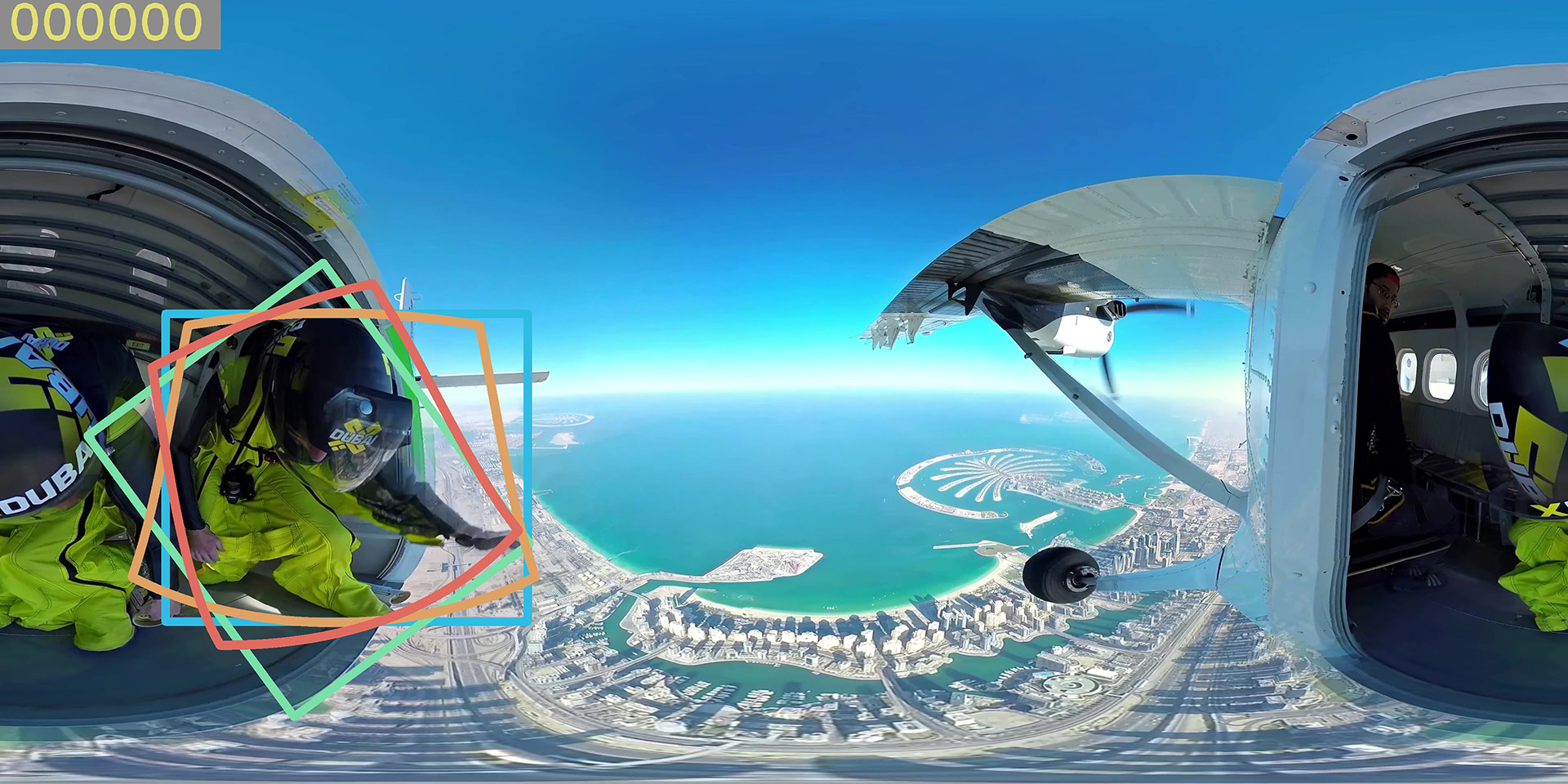}}\!
    \subfloat{\includegraphics[width=\imgw\linewidth, height=\imgh\linewidth]{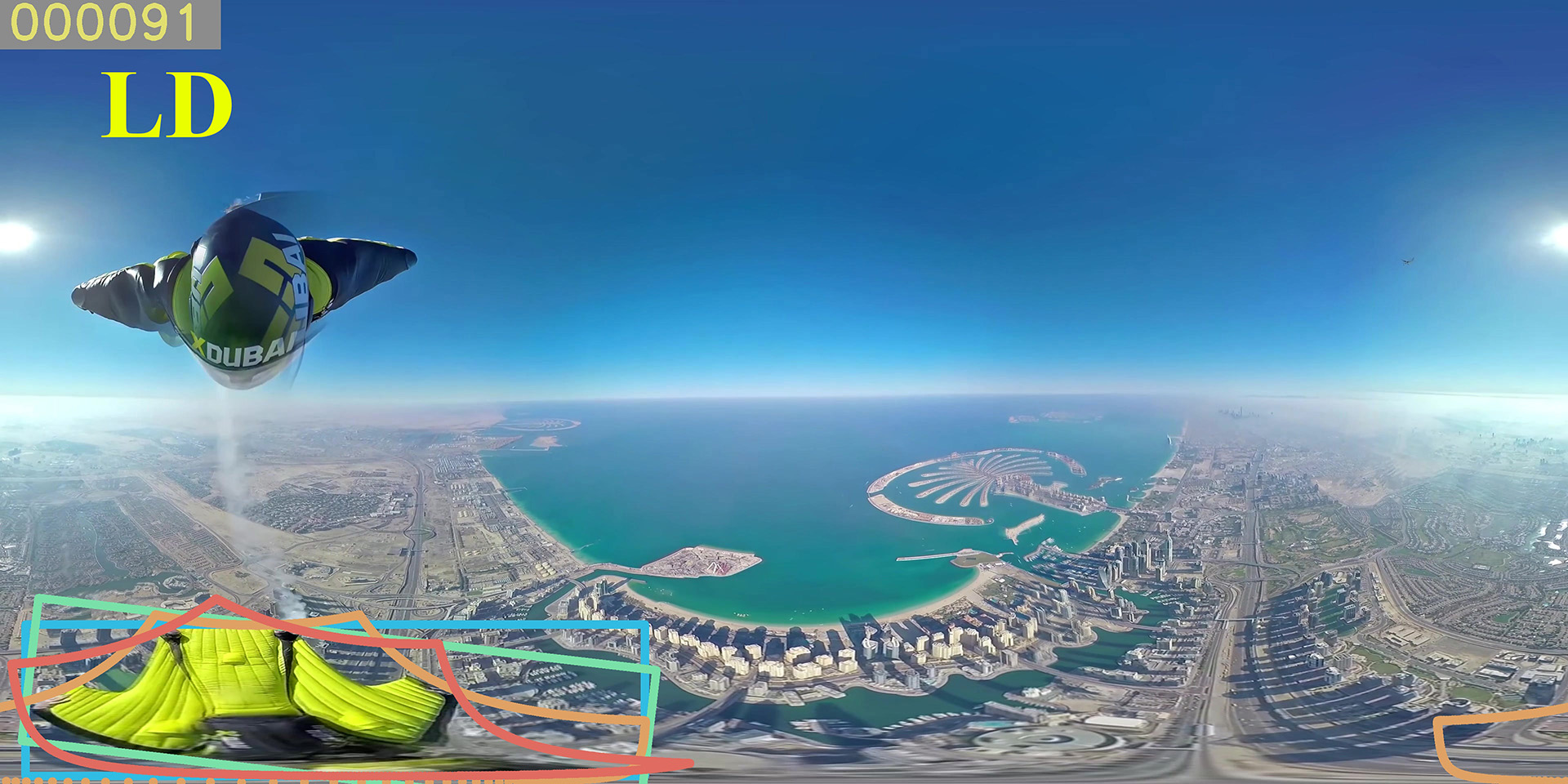}}\!
    \subfloat{\includegraphics[width=\imgw\linewidth, height=\imgh\linewidth]{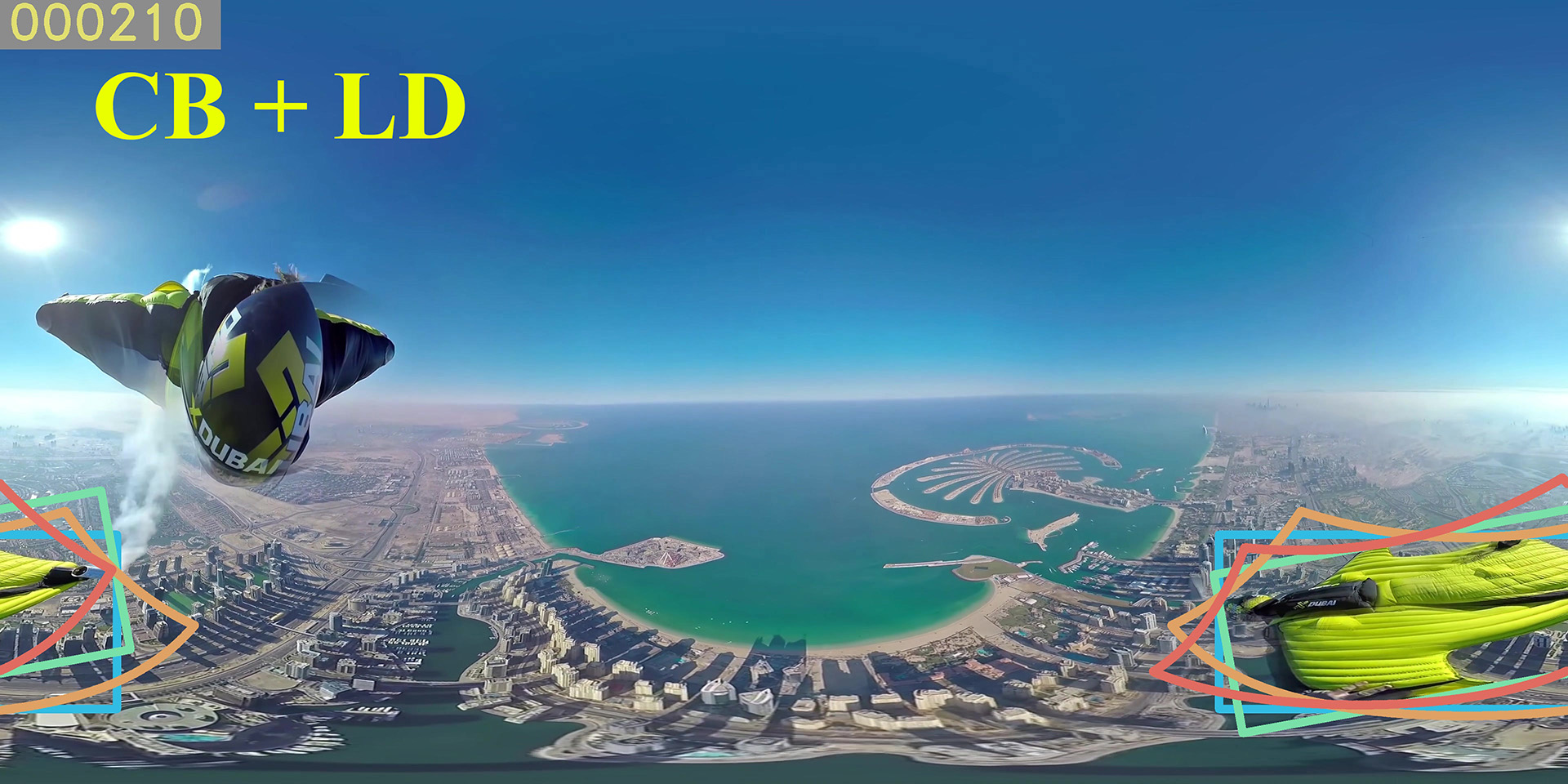}}\!
    \subfloat{\includegraphics[width=\imgw\linewidth, height=\imgh\linewidth]{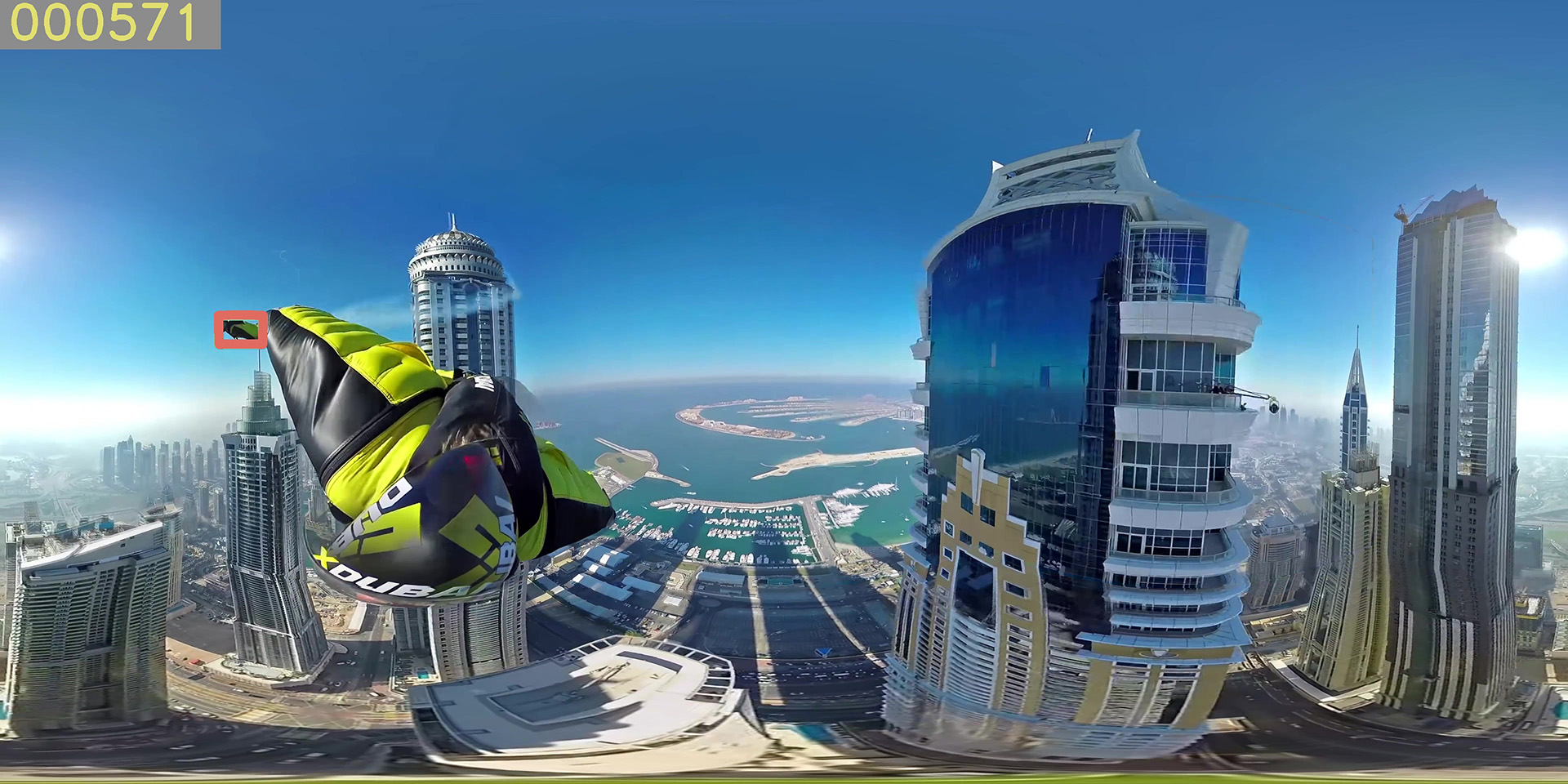}}\\
    \vspace{0.05em}

    \subfloat{\includegraphics[width=\imgw\linewidth, height=\imgh\linewidth]{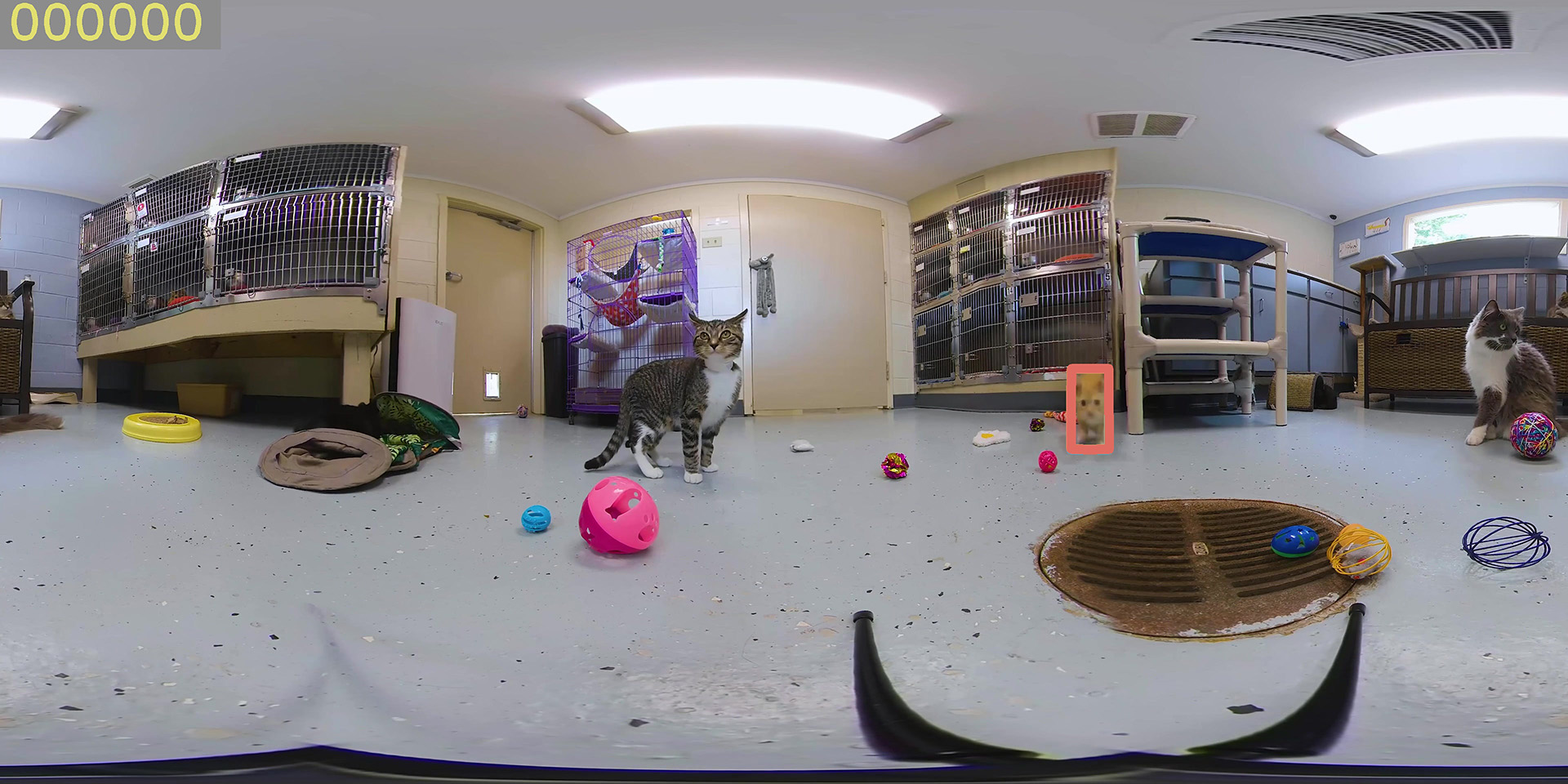}}\!
    \subfloat{\includegraphics[width=\imgw\linewidth, height=\imgh\linewidth]{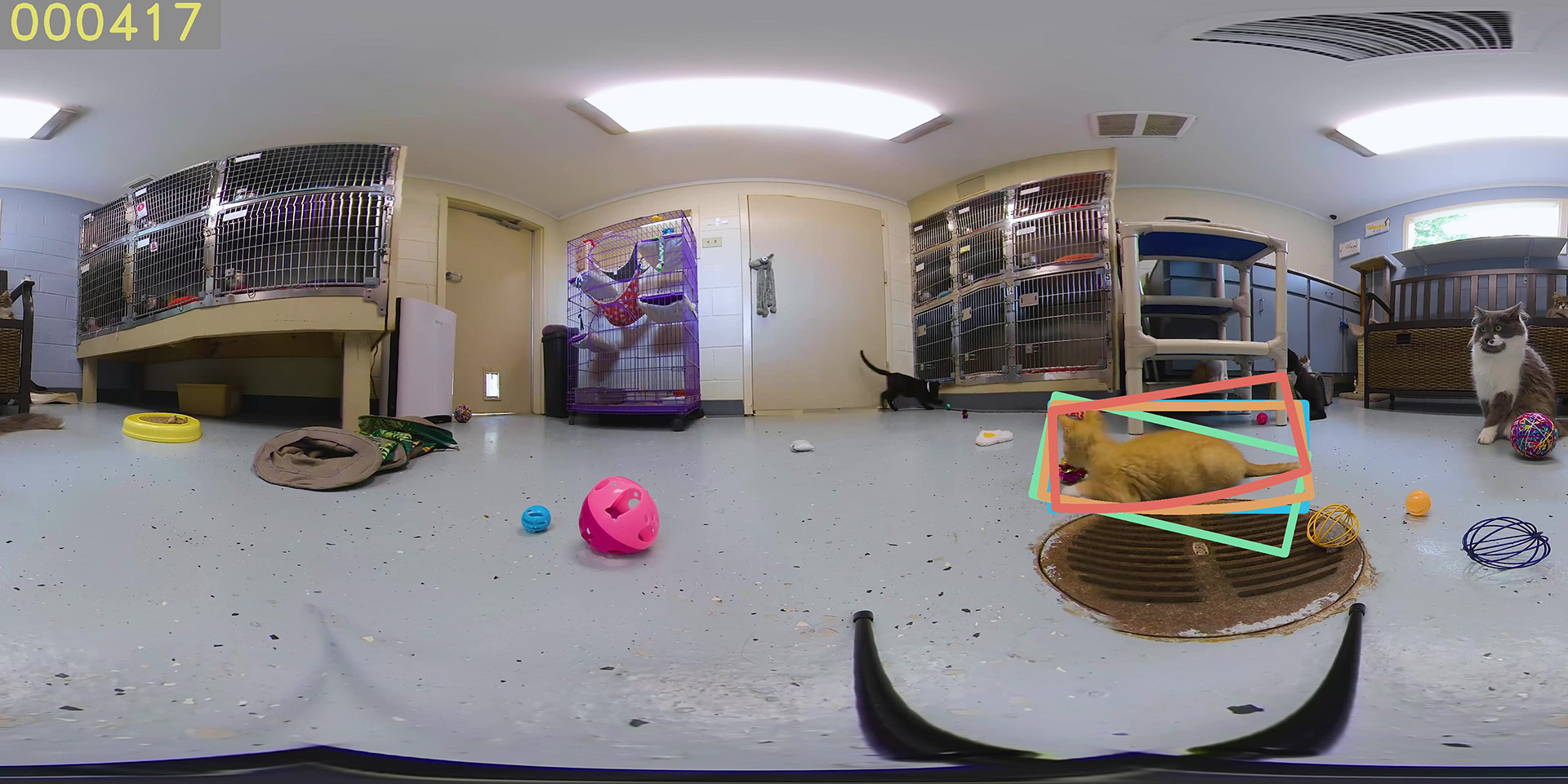}}\!
    \subfloat{\includegraphics[width=\imgw\linewidth, height=\imgh\linewidth]{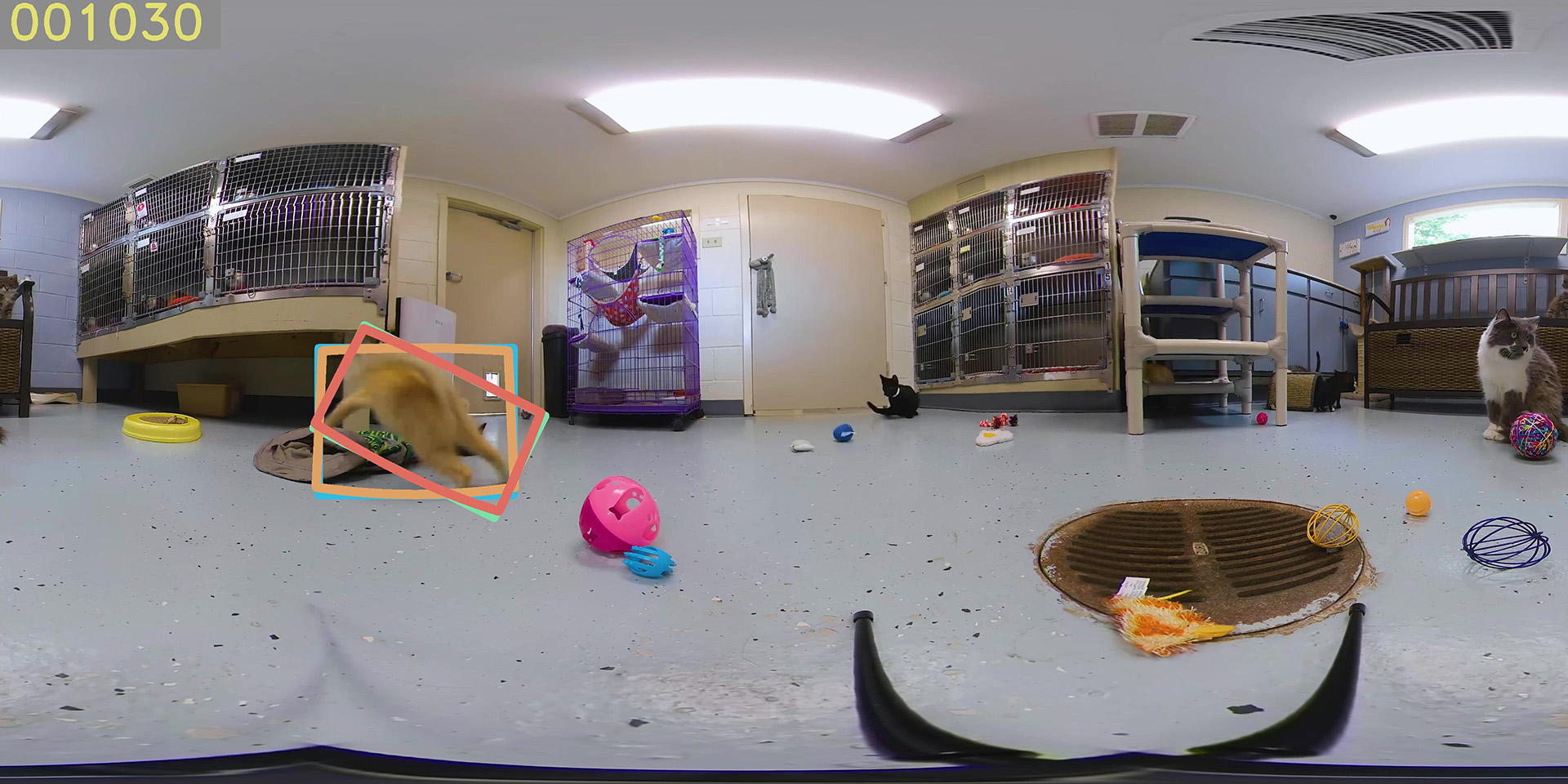}}\!
    \subfloat{\includegraphics[width=\imgw\linewidth, height=\imgh\linewidth]{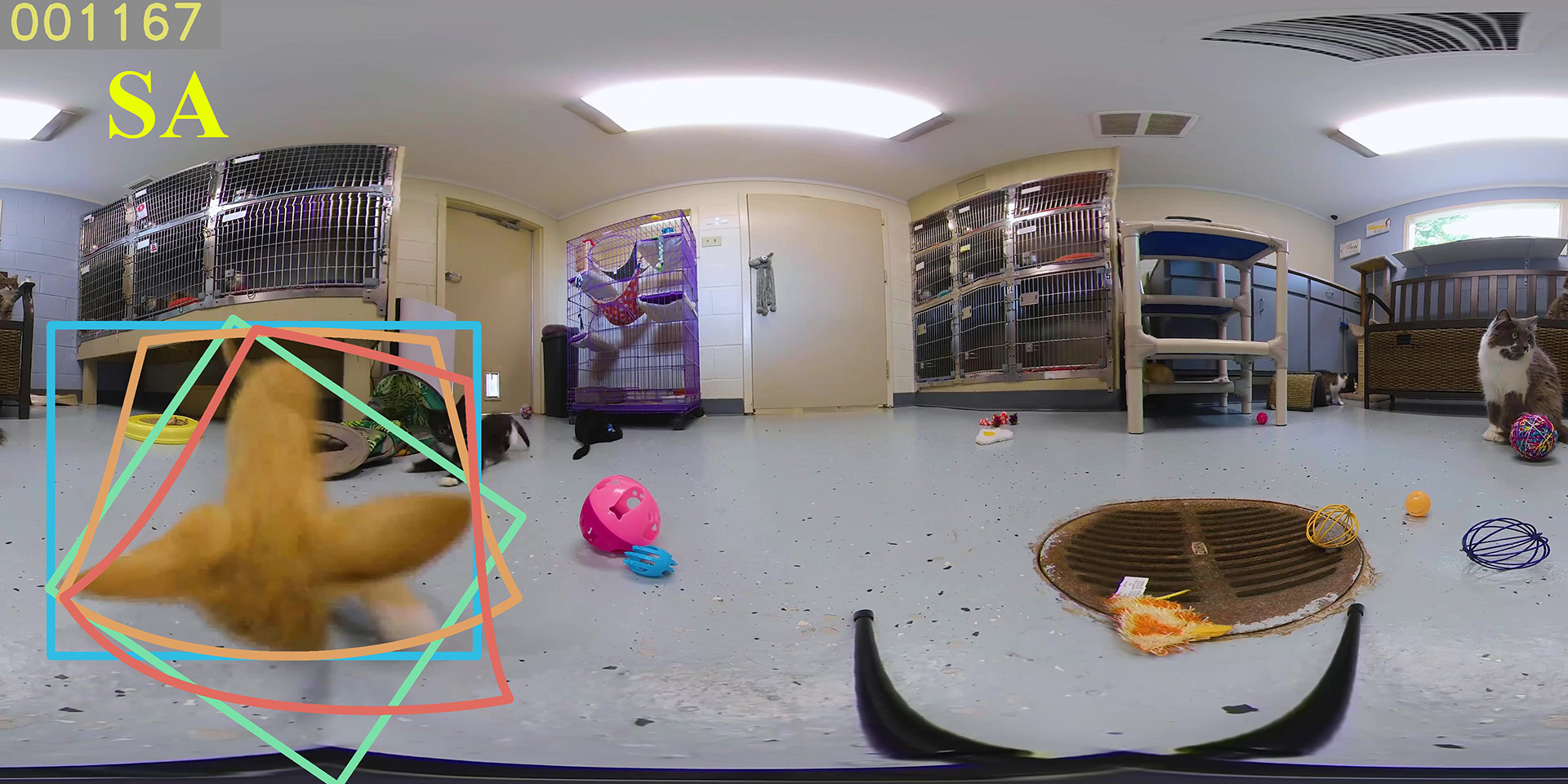}}\\
    \vspace{-0.5em} 
    \caption{Example sequences and annotations of 360VOT benchmark dataset. The target objects in each 360$\degree$ frame are annotated with four different representations as ground truth, including \textcolor{bbox}{bounding box}, \textcolor{rbbox}{rotated bounding box}, \textcolor{bfov}{bounding field-of-view}, and \textcolor{rbfov}{rotated bounding field-of-view.} 360VOT brings distinct challenges for tracking, e.g., crossing border (CB), large distortion (LD) and stitching artifact (SA).  }\label{fig:teaser}
\end{center}%
}]

\ificcvfinal\thispagestyle{empty}\fi

\begin{abstract}
\vskip -0.3cm 
   360$\degree$ images can provide an omnidirectional field of view which is important for stable and long-term scene perception.
   In this paper, we explore 360$\degree$ images for visual object tracking and perceive new challenges caused by large distortion, stitching artifacts, and other unique attributes of 360$\degree$ images. To alleviate these problems, we take advantage of novel representations of target localization, i.e., bounding field-of-view, and then introduce a general 360 tracking framework that can adopt typical trackers for omnidirectional tracking. More importantly, we propose a new large-scale omnidirectional tracking benchmark dataset, 360VOT, in order to facilitate future research. 360VOT contains 120 sequences with up to 113K high-resolution frames in equirectangular projection. The tracking targets cover 32 categories in diverse scenarios. Moreover, we provide 4 types of unbiased ground truth, including (rotated) bounding boxes and (rotated) bounding field-of-views, as well as new metrics tailored for 360$\degree$ images which allow for the accurate evaluation of omnidirectional tracking performance. Finally, we extensively evaluated 20 state-of-the-art visual trackers and provided a new baseline for future comparisons. Homepage: {\small \url{https://360vot.hkustvgd.com}}
   \vskip -0.3cm
\end{abstract}


\section{Introduction}
Visual object tracking is an essential problem in computer vision since it is demanded in various applications such as video analysis, human-machine interaction, and intelligent robots. In the last decade, a large number of visual tracking algorithms~\cite{kcf, mdnet, eco, siamrpn++, dimp} and various benchmarks~\cite{otb100, uav123, VOT, lasot, got10k} have been proposed to promote the development of the visual tracking community. Whereas most existing research focuses on perspective visual object tracking, there is little attention paid to omnidirectional visual object tracking.

Omnidirectional visual object tracking employs a 360$\degree$ camera to track the target object.
With its omnidirectional field-of-view (FoV), a 360$\degree$ camera offers continuous observation of the target over a longer period, minimizing the out-of-view issue. This advantage is crucial for intelligent agents to achieve stable, long-term tracking, and perception.
In general, an ideal spherical camera model is used to describe the projection relationship of a 360$\degree$ camera. The 360$\degree$ image is widely represented by equirectangular projection (ERP)~\cite{erp}, which has two main features: 1) crossing the image border and 2) extreme distortion as the latitude increases. 
Moreover, due to inherent limitations or manufacturing defects of the camera, the 360$\degree$ image may suffer from stitching artifacts that would blur, break or duplicate the shape of objects. Meanwhile, omnidirectional FoV means it is inevitable to capture the photographers. They would distract and occlude the targets. These phenomena are illustrated in Figure~\ref{fig:teaser}. Eventually, they bring new challenges for object tracking on 360$\degree$ images. 

To explore this problem and understand how the current tracking algorithms designed for perspective visual tracking perform, we proposed a challenging omnidirectional tracking benchmark dataset, referred to as 360VOT. The benchmark dataset is composed of 120 sequences, and each sequence has an average of 940 frames with $3840\times1920$ resolution. Our benchmark encompasses a wide range of categories and diverse scenarios, such as indoor, underwater, skydiving, and racing. Apart from 13 conventional attributes, 360VOT has additional 7 attributes, including the aforementioned challenges, fast motion on the sphere and latitude variation. Additionally, we introduce new representations to object tracking. Compared to the commonly used bounding box (BBox), bounding field-of-view (BFoV)~\cite{uiou, PANDORA} represents object localization on the unit sphere in an angular fashion. BFoV can better constrain the target on 360$\degree$ images and is not subject to image resolution. Based on BFoV, we can properly crop the search regions, which enhances the performance of the conventional trackers devised for perspective visual tracking in omnidirectional tracking. To encourage future research, we provide densely unbiased annotations as ground truth, including BBox and three advanced representations, i.e., rotated BBox (rBBox), BFoV, and rotated BFoV (rBFoV). Accordingly, we develop new metrics tailored for 360$\degree$ images to accurately evaluate omnidirectional tracking performances.

In summary, the contribution of this work includes:
\begin{itemize}[topsep=0pt, itemsep=0pt, leftmargin=*]
    \item The proposed 360VOT, to the best of our knowledge, is the first benchmark dataset for omnidirectional visual object tracking. 
    \item We explore the new representations for visual object tracking and provide four types of unbiased ground truth. 
    \item We propose new metrics for omnidirectional tracking evaluation, which measure the dual success rate and angle precision on the sphere.
    \item We benchmark 20 state-of-the-art trackers on 360VOT with extensive evaluations and develop a new baseline for future comparisons.    
\end{itemize}


\begin{table}
    \centering
    \captionsetup{font={color=magenta}}
    \centering
    \tabcolsep=2.5pt
    \footnotesize
   \resizebox{\linewidth}{!}{ 
	\begin{tabular}{r||cc HHHH cccc H} 
    \hline
	Benchmark & Videos & \makecell{Total\\frames} &\makecell{Min\\frames} & \makecell{Mean\\frames} & \makecell{Median\\frames} & \makecell{Max\\frames} & \makecell{Object\\classes} &Attr. &Annotation &Feature &Year\\
    \hline \hline
    {ALOV300}\cite{alov300}&314&152K&19&483&276&5,975&64&14&{sparse BBox}&diverse scenes&2013\\
    {OTB100}\cite{otb100}&100&81K&71&590&393&3,872&16&11&{dense BBox}&short-term&2015\\
    {NUS-PRO}\cite{nus-pro}&365&135K&146&371&300&5,040&8&12&{dense BBox}&occlusion-level&2015\\
    {TC128}\cite{tc128}&129&55K&71&429&365&3,872&27&11&{dense BBox}&{color enhanced}&2015\\
    {UAV123}\cite{uav123}&123&113K&109&915&882&3,085&9&12&{dense BBox}&UAV&2016\\
    {DTB70}\cite{dtb70}&70&16K&68&225&202&699&29&11&{dense BBox}&UAV&2016\\
    {NfS}\cite{nfs}&100&383K&169&3,830&2,448&20,665&17&9&{dense BBox}&high FPS&2017\\
    {UAVDT}\cite{uavdt}&100&78K&82&778&602&2,969&27&14&{sparse BBox}&UAV&2017\\
    
    {TrackingNet}$^*$\cite{trackingnet}&511&226K&96&441&390&2,368&27&15&{sparse BBox}&large scale&2018\\
    {OxUvA}\cite{OxUvA}&337&1.55M&900&4,260&2,628&37,440&22&6&{sparse BBox}&long-term&2018\\
    
    {LaSOT}$^*$\cite{lasot}&280&685K&1,000&2,448&2,102&9,999&85&14&{dense BBox}&{category balance}&2018\\
    {GOT-10k}$^*$\cite{got10k}&420&56K&29&127&100&920&84&6&{dense BBox}&generic&2019\\
    {TOTB}\cite{totb}&225&86K&126&381&389&500&15&12&{dense BBox} &transparent &2021\\
    {TREK-150}\cite{trek150}&150&97K&161&649&484&4,640&34&17&{dense BBox}&FPV&2021\\
    {VOT}\cite{VOT}&62&20K&41&321&242&1,500&37&9&{dense BBox}&annual&2022\\
    \hline 
    {360VOT}&120&113K&251&940&775&2,400&32&20&\makecell{dense (r)BBox\\\& (r)BFoV}&360$\degree$ images &2023\\
    \hline
    \end{tabular}
    }
    \caption{Comparison of current popular benchmarks for visual single object tracking in the literature. $^*$ indicates that only the test set of each dataset is reported. 
    }
    \label{tab:benchmark}
\end{table}
\section{Related work}
\subsection{Benchmarks for visual object tracking}
With the remarkable development of the visual object tracking community, previous works have proposed numerous benchmarks in various scenarios. 
ALOV300~\cite{alov300} is a sparse benchmark introducing 152K frames and 16K annotations, while UAVDT~\cite{uavdt} focuses on UAV scenarios and has 100 videos. TrackingNet~\cite{trackingnet} is a large-scale dataset collecting more than 14M frames based on the YT-BB dataset~\cite{YT-BB}. As YT-BB only provides fine-grained annotations at 1 fps, they explored a tracker to densify the annotations without further manual refinement. OxUvA~\cite{OxUvA} targets long-term tracking by constructing 337 video sequences, but each video only has 30 frames annotated.
One of the first dense BBox benchmarks is OTB100~\cite{otb100} which is extended from OTB50~\cite{otb50} and has 100 sequences. NUS-PRO~\cite{nus-pro} takes the feature of occlusion-level annotation and provides 365 sequences, while TC128~\cite{tc128} researches the chromatic information in visual tracking. 
UAV123~\cite{uav123} and DTB70~\cite{dtb70} offer 123 and 70 aerial videos of rigid objects and humans in various scenes. NfS~\cite{nfs} consists of more than 380K frames captured at 240 FPS studying higher frame rate tracking, while LaSOT~\cite{lasot} is a large-scale and category balance benchmark of premium quality. GOT-10k~\cite{got10k} provides about 1.5M annotations and 84 classes of objects, aiming at generic object tracking. The annual tracking challenge VOT~\cite{VOT} offered 62 sequences and 20K frames in 2022. A more recent benchmark TOTB~\cite{totb} mainly focuses on transparent object tracking. TREK-150~\cite{trek150} introduces 150 sequences of tracking in First Person Vision (FPV) with the 
interaction between the person and the target object.
By contrast, our proposed 360VOT is the first benchmark dataset to focus on object tracking and explore new representations on omnidirectional videos. A summarized comparison with existing benchmarks is reported in Table~\ref{tab:benchmark}.


\subsection{Benchmarks for 360$\degree$ object detection}
Most visual trackers rely on the approaches of tracking by detection. Benefiting from the rapid development of object detection, it is effective to improve the performance of tracking by utilizing those sophisticated network architectures to obtain more robust correlation features. 
Recently, aiming at omnidirectional understanding and perception, researchers started resorting to object detection algorithms for 360$\degree$ images or videos. Several 360$\degree$ datasets and benchmarks for object detection have been proposed. FlyingCars~\cite{FlyingCars} is a synthetic dataset composed of \edited{6K images in 512 $\times$ 256} of synthetic cars and panoramic backgrounds. OSV~\cite{osv2019} created a dataset that covers object annotations on 600 street-view panoramic images. 360-Indoor~\cite{indoor360} focuses on indoor object detection among 37 categories, while PANDORA~\cite{PANDORA} provides 3K images of 1920 × 960 resolution with  rBFoV annotation.
\edited{These 360 detection benchmarks contain independent images with a sole type of annotation. Differently, as a benchmark for visual object tracking, 360VOT contains large-scale 360$\degree$ videos with long footage, higher resolution, diverse environments, and 4 types of annotations. }

\subsection{Visual object tracking scheme}
To guarantee high tracking speed, the trackers for single object tracking generally crop the image and search for the target in small local regions. The tracking scheme is vital in selecting searching regions and interpreting network predictions over sequences in the inference phase. A compatible tracking inference scheme can enhance tracking performance. For example, DaSiamRPN~\cite{desiamrpn} explored a local-to-global searching strategy for long-term tracking. SiamX~\cite{siamx2022} proposed an adaptive inference scheme to prevent tracking loss and realize fast target re-localization. Here, we introduce a 360-tracking framework to make use of local visual trackers, which are trained on normal perspective images to achieve enhanced performance on 360$\degree$ video tracking.

\section{Tracking on 360$\degree$ video}
The 360$\degree$ video is composed of frames using the most common ERP. Each frame can capture 360$\degree$ horizontal and 180$\degree$ vertical field of view. Although omnidirectional FoV avoids out-of-view issues, the target may cross the left and right borders of a 2D image. Additionally, nonlinear projection distortion makes the target largely distorted when they are near the top or bottom of the image, as illustrated in Figure~\ref{fig:teaser}. Therefore, a new representation and framework that fit ERP for 360$\degree$ visual tracking are necessary.

\begin{figure}
    \centering
    \includegraphics[width=\linewidth]{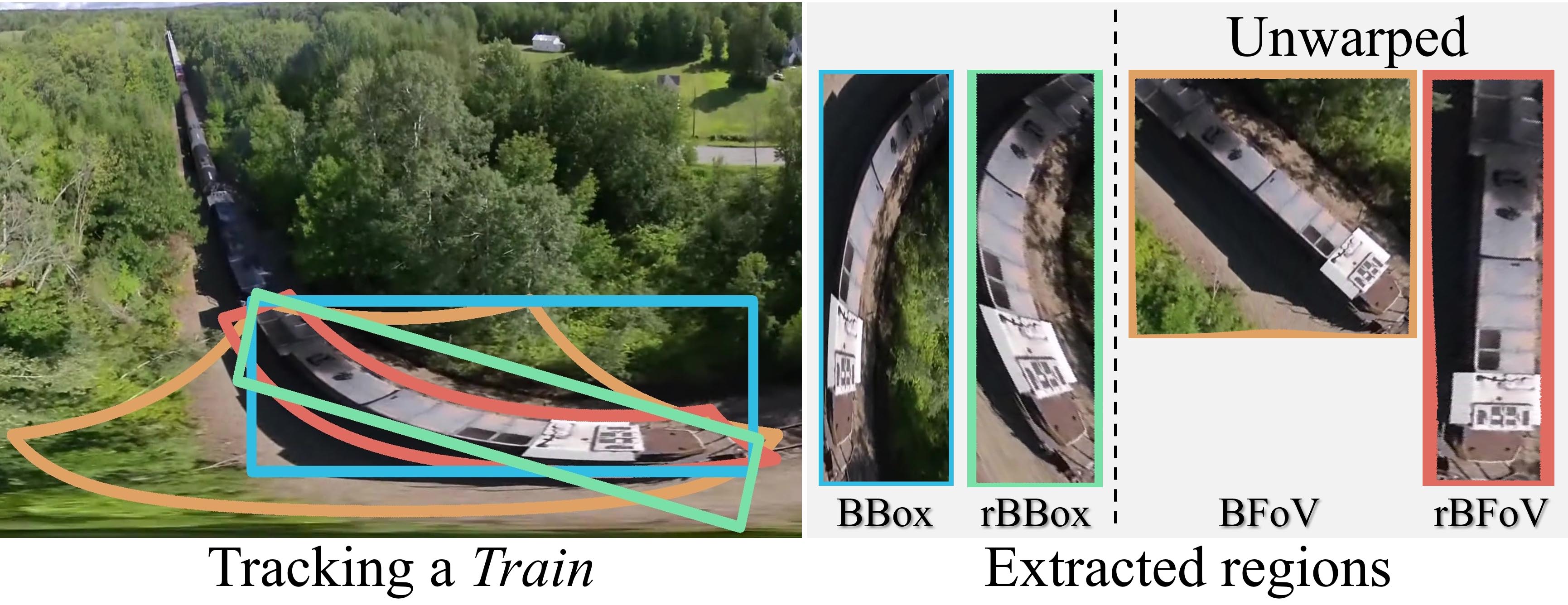}
    \caption{\edited{\textit{Train}}. The comparison of the bounding regions of different representations on a 360$\degree$ image. The \edited{unwarped} images based on BFoV and rBFoV are less distorted.}
    \label{fig:representation}
\end{figure}
 \begin{figure*}
    \includegraphics[width=1\linewidth]{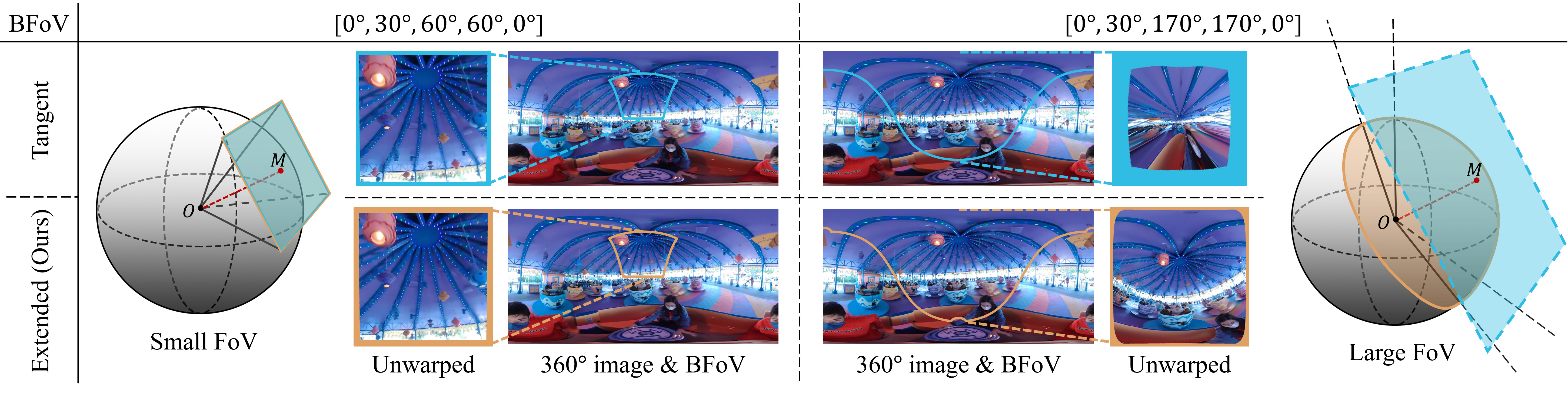}
    \caption{The boundaries on the 360$\degree$ images and the corresponding \edited{unwarped} images of different BFoV definitions. The tangent BFoV is displayed in \textcolor{cyan}{blue} and the extended BFoV is in \textcolor{orange}{orange}. \edited{$M$ on the sphere surface denotes the object center and tangent point. \textcolor{cyan}{Blue} plane with dotted borders represents a larger plane out of space.} Best viewed in color.}
    \label{fig:bfov}
\end{figure*}

\begin{figure*}[!ht]
    \centering
    \includegraphics[width=1\linewidth]{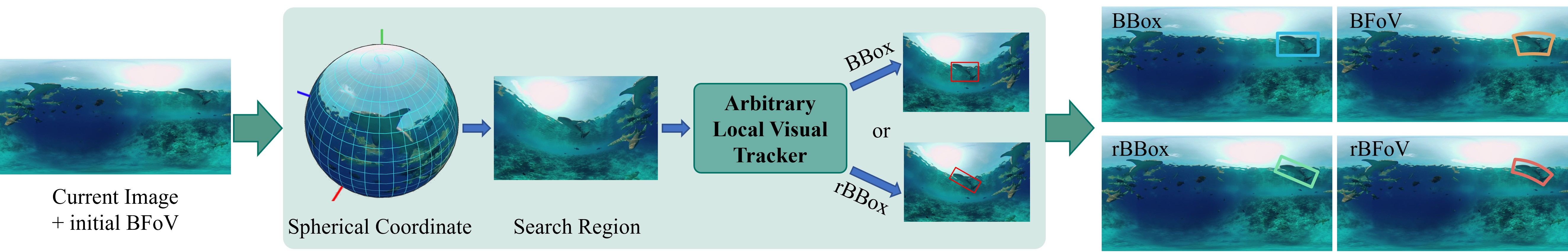}
    \caption{The diagram of 360 tracking framework. 360 tracking framework takes responsibility to extract local search regions for tracking and interpret tracking results. It supports various local visual trackers and can generate 4 types of representation.} 
    \label{fig:360tracking}
\end{figure*} 

\subsection{Representation for the target location}\label{seq:representation}
{The (r)BBox is the most common and simple way to represent the target object's position in perspective images. It is a rectangular area defined by the rectangle around the target object on the image and denoted as $[cx, cy, w, h, \gamma]$, where $cx, cy$ are object center, $w, h$ are width and height. The rotation angle $\gamma$ of BBox is always zero. } However, these representations would become less accurate to properly constrain the target on the 360$\degree$ image. The works~\cite{spherical_criteria, PANDORA} for 360$\degree$ object detection show that the BFoV and rBFoV are more appropriate representations on 360$\degree$ images. Basically, we can use the spherical camera model $\mathcal{F}$ to formulate the mathematical relationship between the 2D image in ERP and a continuous 3D unit sphere~\cite{360vo}. 
(r)BFoV is then defined as $[clon, clat, \theta, \phi, \gamma]$ where $[clon, clat]$ are the longitude and latitude coordinates of the object center at the spherical coordinate system, $\theta$ and $\phi$ denote the maximum horizontal and vertical field-of-view angles of the object’s occupation. 
Additionally, the represented region of (r)BFoV on the 360$\degree$ image is commonly calculated via a tangent plane ~\cite{spherical_criteria, PANDORA} \edited{, $T(\theta, \phi)\in\mathbb{R}^3$, and formulated as:
\begin{equation}
    \small
    I(\mbox{\footnotesize(r)BFoV} \,|\, \Omega) = \mathcal{F}(\mathcal{R}_y(clon)\cdot\mathcal{R}_x(clat)\cdot\mathcal{R}_z(\gamma)\cdot \Omega) ,
\end{equation}
where $\mathcal{R}$ denotes the 3D rotation along the $y,x,z$ axis, $\Omega$ equals $T(\theta, \phi)$ here.}
The \edited{unwarped} images based on tangent BFoV are distortion-free under the small FoV, as shown in Figure \ref{fig:representation}.

However, this definition has a disadvantage on large FoV and cannot represent the region exceeding 180$\degree$ FoV essentially. With the increasing FoV, the \edited{unwarped} images from the tangent planes have drastic distortions, shown in the upper row in Figure~\ref{fig:bfov}. This defect limits the application of BFoV on visual object tracking since trackers rely on \edited{unwarped} images for target searching.  
To address this problem, we extended the definition of BFoV. When the bounding region involves a large FoV, i.e., larger than 90$\degree$, 
\edited{the extended BFoV leverages a spherical surface  $S(\theta, \phi)\in\mathbb{R}^3$ instead of a tangent plane to represent the bounding region on the 360$\degree$ image.
Therefore, the corresponding region of extended (r)BFoV on 360$\degree$ is formulated as:}
\edited{\begin{equation}
    \small
    I(\mbox{\footnotesize(r)BFoV} \,|\, \Omega), \quad  \Omega = \begin{cases}
    T(\theta, \phi), &\theta<90\degree, \phi<90\degree\\
    S(\theta, \phi), &otherwise
    \end{cases} .\label{eq:ebfov}
\end{equation} }
The comparisons of the boundary on 360$\degree$ images and corresponding \edited{unwarped} images based on tangent BFoV and the extended BFoV are shown in Figure ~\ref{fig:bfov}. \edited{Please refer to the supplementary for the detailed formulation.}

\subsection{360 tracking framework}
\edited{To conduct omnidirectional tracking using an existing visual tracker, we propose a 360 tracking framework, shown in Figure~\ref{fig:360tracking}.} The framework leverages extended BFoV to address challenges caused by object crossing-border and large distortion on the 360$\degree$ image. As a continuous representation in the spherical coordinate system, BFoV is not subject to the resolution of the image. \edited{Given an initial BFoV, the framework first calculates the corresponding region $I$ on the 360$\degree$ image via Eq~\ref{eq:ebfov}.
By remapping the 360$\degree$ image using pixel coordinates recorded in $I$, it extracts a less distorted local search region for target identification. From this extracted image, a local visual tracker then infers a BBox or rBBox prediction. 
Finally, we can still utilize $I$ to convert the local prediction back to the global bounding region on the 360$\degree$ image. The (r)BBox prediction is calculated as the minimum area (rotated) rectangle on the 360$\degree$ image. In terms of (r)BFoV, we can re-project the bounding region's coordinates onto the spherical coordinate system and calculate the maximum bounding FoV. Since the framework does not rely on nor affect the network architecture of the tracker}, we can adapt an arbitrary local visual tracker trained on conventional perspective images for omnidirectional tracking. 



\section{A New Benchmark Dataset: 360VOT}
In this section, we elaborate on how to collect appropriate 360$\degree$ videos and efficiently obtain unbiased ground truth, making a new benchmark dataset for omnidirectional (360$\degree$) Visual Object Tracking (360VOT). 


\subsection{Collection}

The resources of 360$\degree$ videos are much less abundant than normal format videos. We spent lots of \edited{effort} and time collecting hundreds of candidate videos from YouTube and \edited{captured} some using a 360$\degree$ camera. After that, we ranked and filtered them considering four criteria of tracking difficulty scale and some additional challenging cases. Videos can gain higher ranking with 1) adequate relative motion of the target, 2) higher variability of the environment, 3) the target crossing frame borders, and 4) sufficient footage.
In addition to the criteria listed above, videos with additional challenges are assigned a higher priority. For example, distinguishing targets from other highly comparable objects is a challenge in object detection and tracking. 

After filtering, videos are further selected and sampled into sequences with a frame number threshold ($\leq$2400). The relatively stationary frames are further discarded manually. Considering the distribution balance, 120 sequences are finally selected as the 360VOT benchmark. 
The object classes mainly cover \textit{humans} (skydiver, rider, pedestrian and diver), \textit{animals} (dog, cat, horse, shark, bird, monkey, dolphin, panda, rabbit, squirrel, turtle, elephant and rhino), \textit{rigid objects} (car, F1 car, bike, motorbike, boat, aircraft, Lego, basket, building, kart, cup, drone, helmet, shoes, tire and train) and \textit{human \& carrier cases} (human \& bike, human \& motorbike and human \& horse). Our benchmark encompasses a wide range of categories with high diversity, as illustrated in the examples in Figure~\ref{fig:teaser}. 

\subsection{Annotation}
Manual annotation of large-scale images in high quality usually requires sufficient manpower with basic professional knowledge in the domain. Accordingly, the tracking benchmark with 4 different types of ground truth increases the manual annotation workload largely increased and makes the annotation standard inconsistent in a large group of annotators. The large distortion and crossing border issues of 360$\degree$ images also make it difficult to obtain satisfactory annotations. Besides, there is no toolkit able to produce BFoV annotation directly. 
To overcome these problems, we seek to segment the per-pixel target instance in each frame and then obtain corresponding optimal (r)BBox and (r)BFoV from the resultant masks.

To realize the objective at a speedy time, our annotation pipeline includes three steps, initial object localization, interactive segmentation refinement, and mask-to-bounding box conversion. First, we integrated our 360 tracking framework with a visual tracker~\cite{siamx2022} and then used it to generate initial BBoxes for all sequences before segmentation. The annotators inspected the tracking results online and would correct and restart the tracking when tracking failed.
The centroid of each BBox would be used to initiate segmentation later.
Second, we developed an efficient segmentation annotation toolkit based on a click-based interactive segmentation model \cite{ritm2022}, which allows annotators to refine the initial segmentation with a few clicks.
Finally, we converted the fine-grained segmentation masks with two rounds of revision to get the four unbiased ground truths by minimizing the bounding areas respectively. Please refer to the supplementary for details of the annotation toolkit and conversion algorithms.

\begin{table}[t]
    \centering\footnotesize
    \tabcolsep=0.06cm 
    \begin{tabular}{c|m{0.85\linewidth}}
    \hline
    Attr. & Meaning\\
    \hline
    IV & The target is subject to light variation.\\ 
    BC & The background has a similar appearance as the target.\\
    DEF & The target deforms during tracking.\\
    MB & The target is blurred due to motion.\\
    CM & The camera has abrupt motion.\\
    ROT & The target rotates related to the frames. \\ 
    POC & The target is partially occluded.\\
    FOC & The target is fully occluded.\\
    ARC & The ratio of the annotation aspect ratio of the first and the current frame is outside the range [0.5, 2].\\
    SV & The ratio of the annotation area of the first and the current frame is outside the range [0.5, 2].\\
    FM & The motion of the target center between contiguous frames exceeds its own size. \\ 
    LR & The area of the target annotation is less than $1000$ pixels.\\ 
    HR & The area of the target annotation is larger than $500^2$ pixels.\\
    \hline\hline
    SA & The 360$\degree$ images have stitching artifacts and they affect the target object.\\
    CB & The target is crossing the border of the frame and partially appears on the other side.\\
    FMS & The motion angle on the spherical surface of the target center is larger than the last BFoV.\\
    LFoV & The vertical or horizontal FoV of the BFoV is larger than $90\degree$.\\
    LV & The range of the latitude of the target center across the video is larger than $50\degree$.\\
    HL & The latitude of the target center is outside the range $[-60\degree, 60\degree]$, lying in the ``frigid zone".\\
    LD & The target suffers large distortion due to the equirectangular projection.\\
    \hline
    \end{tabular}
    \caption{Attribute description. The 360VOT not only contains 13 attributes widely used by the existing benchmarks but also has 7 additional attributes, described in the last block of the row, leading to distinct challenges.}
    \label{tab:attibute}
\end{table}

\begin{figure}[t]
    \centering
    \captionsetup[subfloat]{skip=0pt}
    \subfloat[Atrribute histogram\label{fig:attr_hist}]{\includegraphics[width=0.9\linewidth]{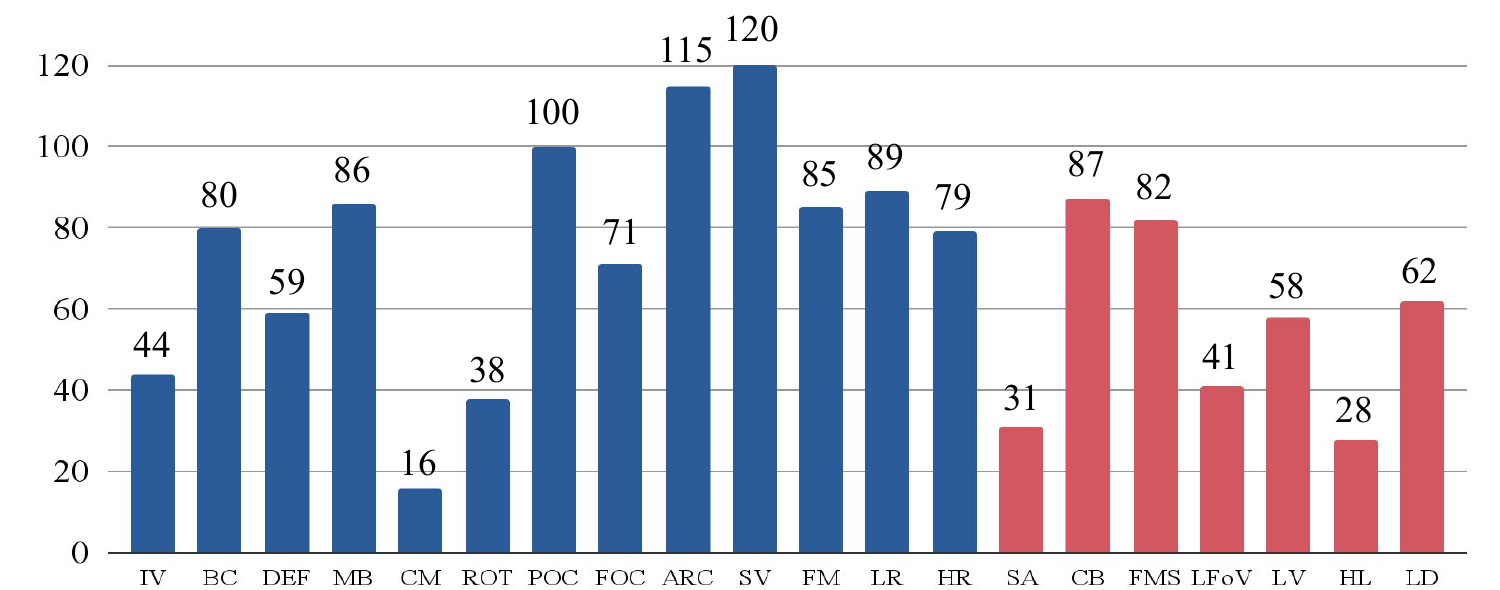}}\\
    \subfloat[Correspondence heatmap\label{fig:attr_heat}]{\includegraphics[width=0.8\linewidth]{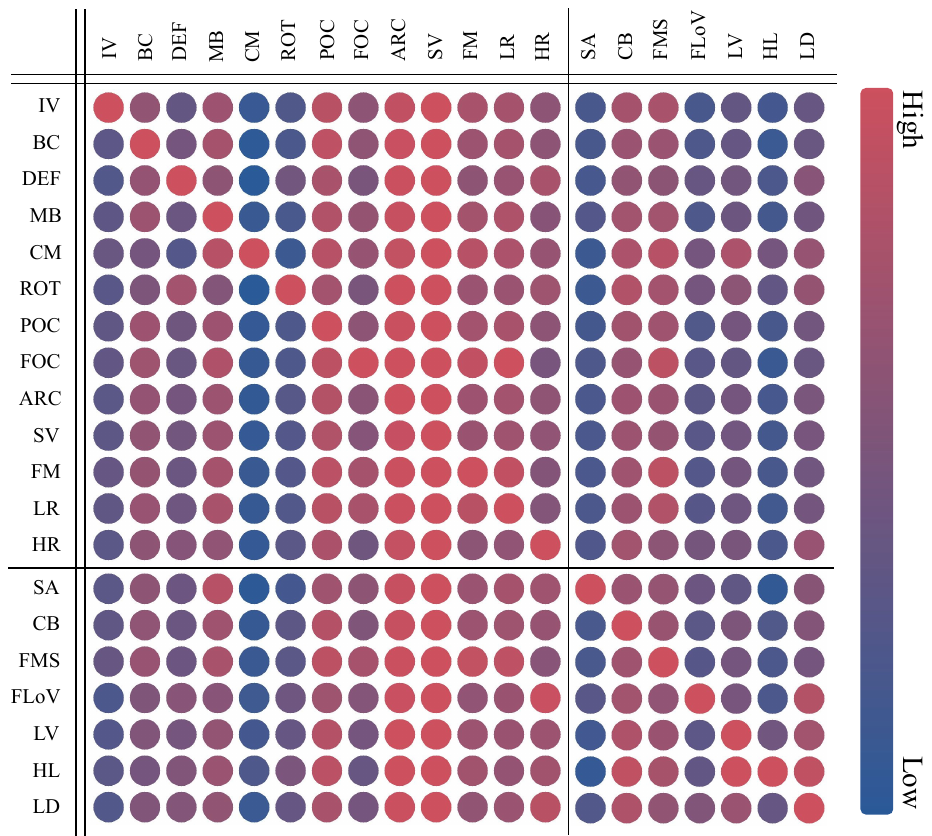}}
    \vspace{-0.5em}
    \caption{Attribute distribution of 360VOT benchmark. } 
    \label{fig:attr_distribution}
\end{figure}

\subsection{Attributes}
Each sequence is annotated with a total of 20 different attributes: illumination variation (IV), background clutter (BC), deformable target (DEF), motion blur (MB), camera motion (CM), rotation (ROT), partial occlusion (POC), full occlusion (FOC), aspect ratio change (ARC), scale variation (SV), fast motion (FM), low resolution (LR), high resolution (HR), stitching artifact (SA), crossing border (CB), fast motion on the sphere (FMS), large FoV (LFoV), latitude variation (LV), high latitude (HL) and large distortion (LD). 
The detailed meaning of each attribute is described in Table~\ref{tab:attibute}. Among them, IV, BC, DEF, MB, CM, ROT, POC and LD attributes are manually labeled, while the others are computed from the annotation results of targets. The distinct features of the 360$\degree$ image are well represented in 360VOT: \textit{location variations} (FMS, LFoV and LV), \textit{external disturbances} (SA and LD) and \textit{special imaging} (CB and HL). See visual examples in the supplementary.

Overall, the exact number of each attribute is plotted in a histogram, as shown in Figure~\ref{fig:attr_hist}, while the correspondence of each attribute is provided with a heatmap, as shown in Figure~\ref{fig:attr_heat}. 
A warmer tone indicates that the pair of attributes are more frequently present together and vice versa. The co-occurrence counts of each row are then normalized by the diagonal counts. 
We observe that \textit{scale changes} (ARC and SV) and \textit{motion} (MB and FM) are common challenges that are also included in other benchmarks.
Extra challenges of 360 tracking, including CB, FMS, LV, and LD, co-occur with traditional challenges.
Specifically, CB occurs when the two patches of the target without intersection are at opposite edges or corners of the frame, and LD happens when the target is of large FoV and appears in a high latitude area of the frame. 


\section{Experiments}
\subsection{Metrics}
To conduct the experiments, we use the standard one-pass evaluation (OPE) protocol~\cite{otb100} and measure the success S, precision P, and normalized precision $\overline{P}$ ~\cite{trackingnet} of the trackers over the 120 sequences. Success S is computed as the intersection over union (IoU) between the tracking results $B^{tr}$ and the ground truth annotations $B^{gt}$. The trackers are ranked by the area under curve (AUC), which is the average of the success rates corresponding to the sampled thresholds $[0, 1]$. The precision P is computed as the distance between the results $\mathbf{C}^{tr}$ and the ground truth centers $\mathbf{C}^{gt}$. The trackers are ranked by the precision rate on the specific threshold (i.e., 20 pixels). The normalized precision $\overline{\mbox{P}}$ is scale-invariant, which normalizes the precision P over the size of the ground truth and then ranks the trackers using the AUC for the $\overline{\mbox{P}}$ between 0 and 0.5. For the perspective image using (r)BBox, these metrics can be formulated as:
    \vspace{-1em}
	\begin{equation}\label{eq:metric1}
		\begin{split}
			\mbox{S}= IoU(B^{gt}, B^{tr}), \,  \mbox{P} = ||\mathbf{C}^{gt}_{xy} - \mathbf{C}^{tr}_{xy}||_2 \\
			\overline{\mbox{P}} = ||diag(B^{gt}, B^{tr})(\mathbf{C}^{gt}_{xy} - \mathbf{C}^{tr}_{xy})||_2 .
		\end{split}
	\end{equation}

However, for 360$\degree$ images, the target predictions may cross the image. To handle this situation and increase the accuracy of BBox evaluation, we introduce dual success S$_{dual}$ and precision P$_{dual}$. Specifically, we shift the $B^{gt}$ to the left and right by $W$, the width of 360$\degree$ images, to obtain two temporary ground truth $B^{gt}_l$ and $B^{gt}_r$. Based on the new ground truth, we then calculate extra success S$_{l}$ and S$_{r}$ and precision P$_{l}$ and P$_{r}$ using Eq.~\ref{eq:metric1}. Finally, S$_{dual}$ and P$_{dual}$ are measured by:
\vspace{-0.5em}
\begin{equation}\label{eq:metric2}
    \begin{split}
    \mbox{S}_{dual} &= max\{\mbox{S}_l, \mbox{S}, \mbox{S}_r\}, \\ \mbox{P}_{dual} &= min\{\mbox{P}_l, \mbox{P}, \mbox{P}_r\} .
    \end{split}
\end{equation}
S$_{dual}$ and S, as P$_{dual}$ and P, are the same when the annotation does not cross the image border. Similarly, we can compute the normalized dual $\overline{\mbox{P}}_{dual}$. 

Since objects suffer significant non-linear distortion in the polar regions due to the equirectangular projection, the distance between the predicted and ground truth centers may be large on the 2D image but they are adjacent on the spherical surface. It means that precision metric P$_{dual}$ is sensitive for 360$\degree$ images. Therefore, we propose a new metric P$_{angle}$, which is measured as the angle precision $<\mathbf{C}^{gt}_{lonlat}, \mathbf{C}^{tr}_{lonlat}> $ between the vectors of the ground truth and the tracker results in the spherical coordinate system. The different trackers are ranked with angle precision rate on a threshold, i.e., 3$\degree$. 
Moreover, when target positions are represented by BFoV or rBFoV, we utilize spherical IoU~\cite{uiou} to compute the success metric, denoted as S$_{sphere}$, while only S$_{sphere}$ and P$_{angle}$ are measured.

\subsection{Baseline trackers}
We evaluated 20 state-of-the-art visual object trackers on 360VOT. According to the latest development of visual tracking, the compared methods can be roughly classified into three groups: transformer trackers, Siamese trackers, and other deep learning based trackers. Specifically, the transformer trackers contain Stark~\cite{stark}, ToMP~\cite{tomp}, MixFormer~\cite{mixformer}, SimTrack~\cite{simtrack} and AiATrack~\cite{aiatrack}. The Siamese trackers include SiamDW~\cite{SiamDW}, SiamMask~\cite{siammask}, SiamRPNpp~\cite{siamrpn++}, SiamBAN~\cite{siamban}, AutoMatch~\cite{automatch}, Ocean~\cite{ocean} and SiamX~\cite{siamx2022}. For other deep trackers,  UDT~\cite{UDT}, Meta-SDNet~\cite{meta}, MDNet~\cite{mdnet}, ECO~\cite{eco}, ATOM~\cite{atom}, KYS~\cite{kys}, DiMP~\cite{dimp}, PrDiMP~\cite{prdimp} are evaluated. We used the official implementation, trained models, and default configurations to ensure a fair comparison among trackers. In addition, we developed a new baseline AiATrack-360 that combines the transformer tracker AiATrack~\cite{aiatrack} with our 360 tracking framework. We also adapt a different kind of tracker SiamX~\cite{siamx2022} with our framework, named SiamX-360, to verify the generality of the proposed framework.

\begin{table}[!t]
    \centering
    \footnotesize
    \begin{tabular}{lcccc}
    \toprule
	\multirow{2}{*}{Tracker}& \multicolumn{4}{c}{360VOT BBox}\\ \cmidrule(lr){2-5}
	&\makecell{$\mbox{S}_{dual}$\\\footnotesize(AUC)} &$\mbox{P}_{dual}$ &\makecell{$\overline{\mbox{P}}_{dual}$\\\footnotesize(AUC)} &$\mbox{P}_{angle}$\\
	\midrule            
    UDT~\cite{UDT}             &0.104  &0.075  &0.117  &0.098\\
    Meta-SDNet~\cite{meta}     &0.131  &0.097  &0.164  &0.136\\
    MDNet~\cite{mdnet}         &0.150  &0.106  &0.188  &0.143\\
    ECO~\cite{eco}             &0.175  &0.130  &0.212  &0.179\\
    ATOM~\cite{atom}           &0.252  &0.216  &0.286  &0.266\\
    KYS~\cite{kys}             &0.286  &0.245  &0.312  &0.296\\
    DiMP~\cite{dimp}           &0.290  &0.247  &0.315  &0.299\\
    PrDiMP~\cite{prdimp}       &\nd{0.341}  &\nd{0.292}  &\nd{0.371}  &\nd{0.347}\\
    \midrule                      
    SiamDW~\cite{SiamDW}       &0.156  &0.116  &0.190  &0.156\\
    SiamMask~\cite{siammask}   &0.189  &0.161  &0.220  &0.203\\
    SiamRPNpp~\cite{siamrpn++} &0.201  &0.175  &0.233  &0.213\\
    SiamBAN~\cite{siamban}     &0.205  &0.187  &0.242  &0.227\\
    AutoMatch~\cite{automatch} &0.208  &0.202  &0.261  &0.248\\
    Ocean~\cite{ocean}         &0.240  &0.223  &0.287  &0.264\\
    SiamX~\cite{siamx2022}     &\nd{0.302}  &\nd{0.265}  &\nd{0.331}  &\nd{0.315}\\
    \midrule
    Stark~\cite{stark}         &0.381  &0.356  &0.403  &0.408\\
    ToMP~\cite{tomp}           &0.393  &0.352  &0.421  &0.413\\
    MixFormer~\cite{mixformer} &0.395  &\nd{0.378}  &0.417  &\nd{0.424}\\
    SimTrack~\cite{simtrack}   &0.400  &0.373  &0.421  &\nd{0.424}\\
    AiATrack~\cite{aiatrack}   &\nd{0.405}  &0.369  &\nd{0.427}  &0.423\\
    \hline & \\[\dimexpr-\normalbaselineskip+3pt]
    SiamX-360       &0.391  &0.365  &0.430  &0.425\\
    AiATrack-360    &\st{0.534}  &\st{0.506}  &\st{0.563}  &\st{0.574}\\
    \bottomrule
    \end{tabular}
    \vspace{-0.2em}
    \caption{Overall performance on 360VOT BBox in terms of dual success, dual precision, normalized dual precision, and angle precision. Bold \nd{blue} indicates the best results in the tracker group. Bold \st{red} indicates the best results overall.}
    \label{tab:result}
\end{table}

\subsection{Performance based on BBox}
\noindent\textbf{Overall performance}. Existing trackers take the BBox of the first frame to initialize the tracking, and the inference results are also in the form of BBox. 
Table \ref{tab:result} shows comparison results among four groups of trackers, i.e., other, Siamese, transformer baselines, and the adapted trackers for each block in the table. According to the quantitative results, PrDiMP~\cite{prdimp}, SiamX~\cite{siamx2022} and AiATrack-360 perform best in their group of trackers. Owing to the powerful network architectures, the \textit{transformer} trackers generally outperform other groups of the compared trackers. After AiATrack integrates our proposed framework, AiATrack-360 achieves a significant performance increase of 12.9\%, 13.7\%, 13.6\% and 15.1\% in terms of S$_{dual}$, P$_{dual}$, $\overline{\mbox{P}}_{dual}$ and P$_{angle}$ respectively. AiATrack-360 outperforms all other trackers with a big performance gap. Compared to SiamX, SiamX-360 is improved by 8.9\% S$_{dual}$, 10\% P$_{dual}$, 9.6\% $\overline{\mbox{P}}_{dual}$ and 11\% P$_{angle}$, which is comparable with other transformer trackers. Although the performance gains of AiATrack-360 and SiamX-360 are different, it validates the effectiveness and generalization of our 360 tracking framework on 360$\degree$ visual object tracking. They can serve as a new baseline for future comparison.

\begin{figure*}
    \captionsetup[subfloat]{skip=0pt}
    \centering
    \def\imw{0.33}
    \captionsetup[subfigure]{labelformat=empty}
    \subfloat{\includegraphics[width=\imw\linewidth]{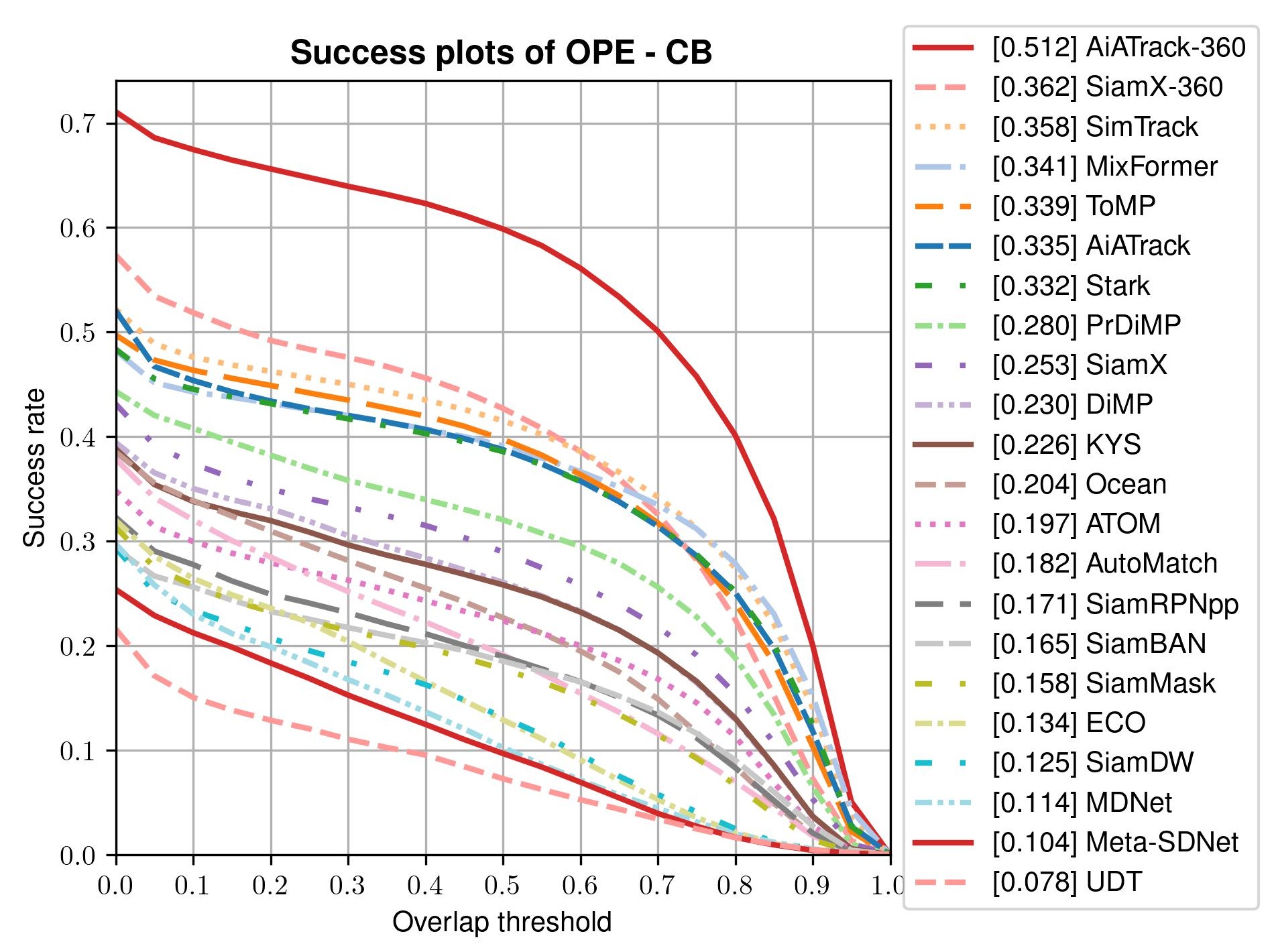}}
    \subfloat{\includegraphics[width=\imw\linewidth]{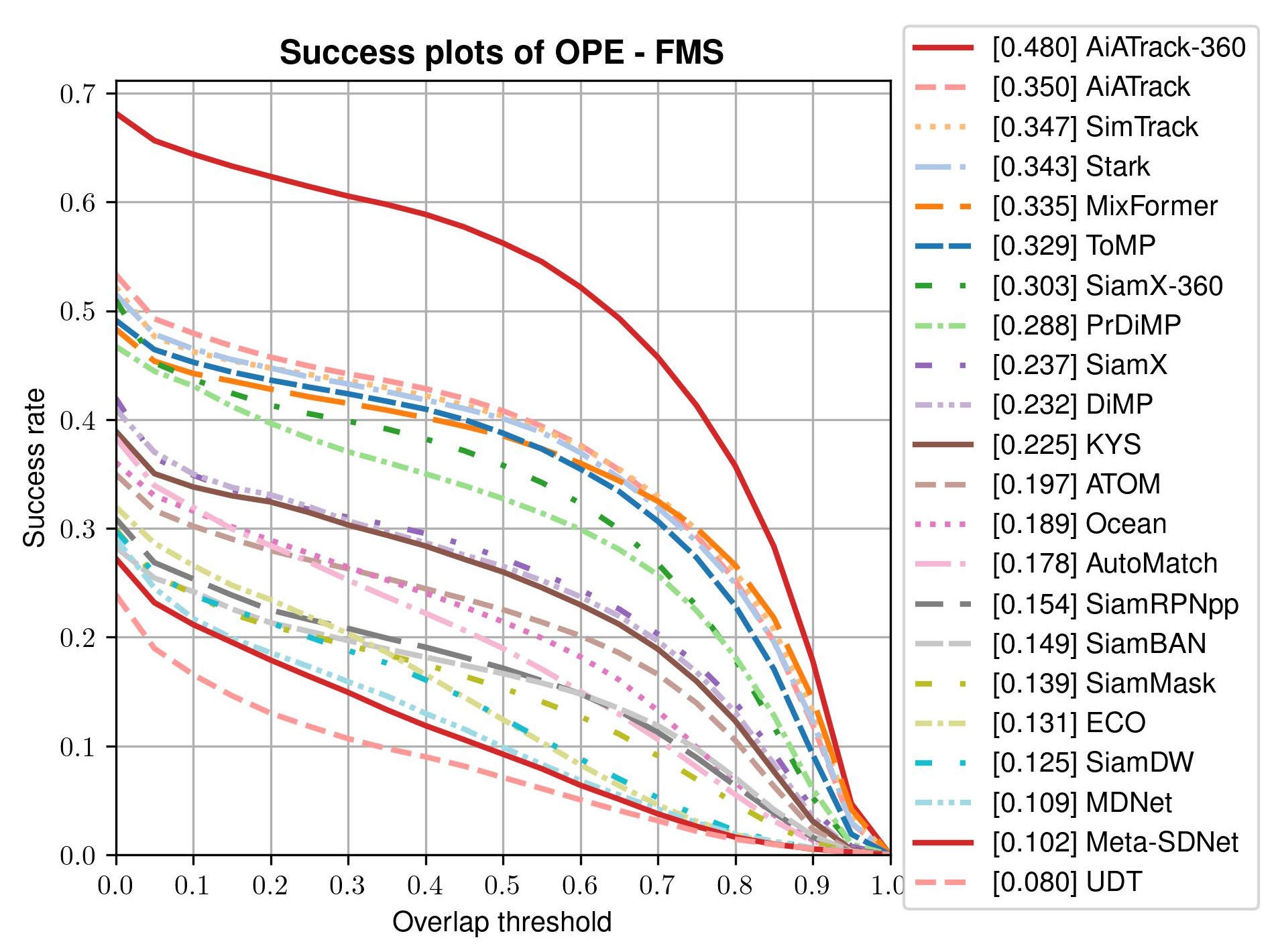}}
    \subfloat{\includegraphics[width=\imw\linewidth]{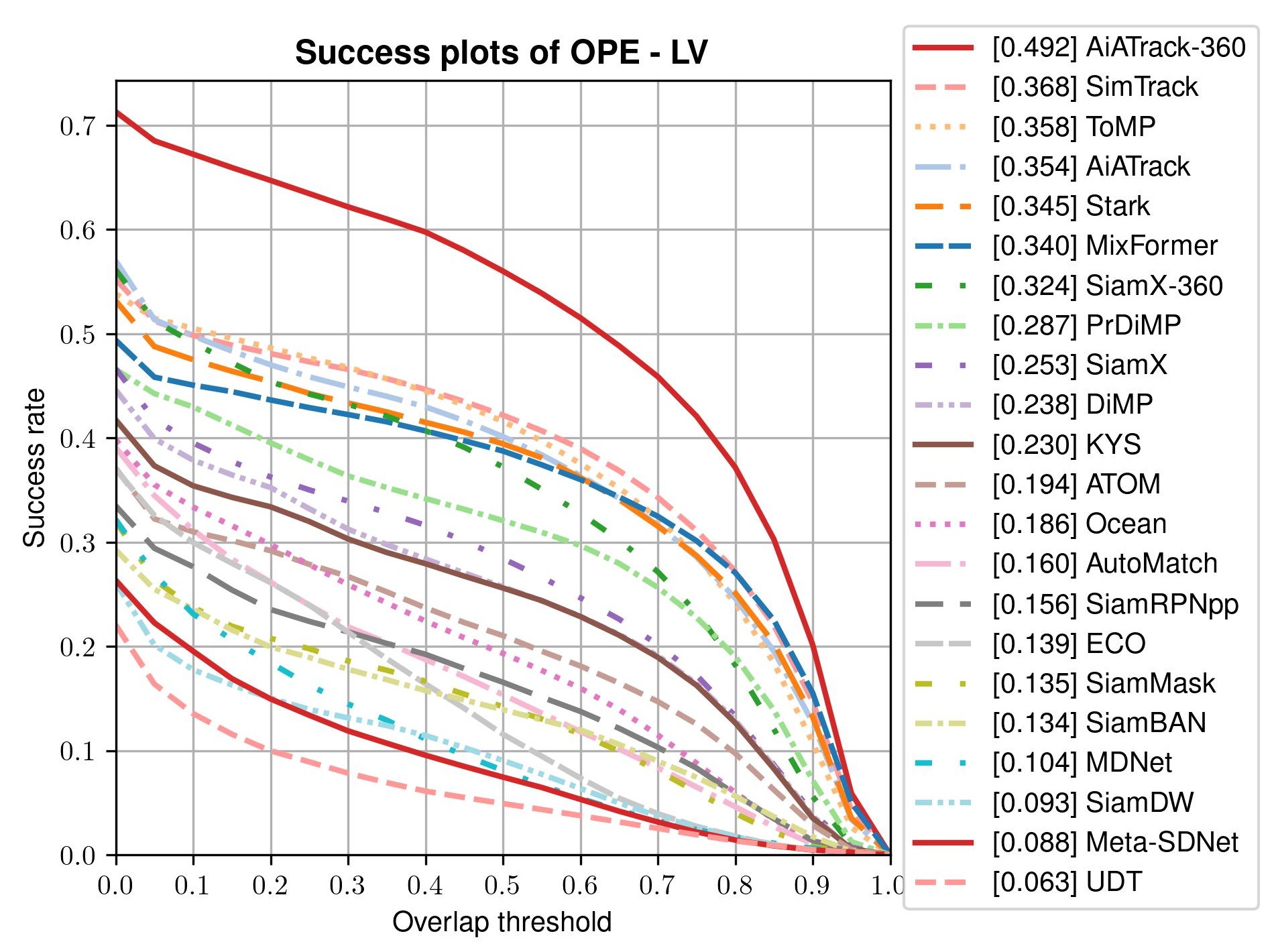}}\\

    \subfloat[\edited{(a) Crossing Border}]{\includegraphics[width=\imw\linewidth]{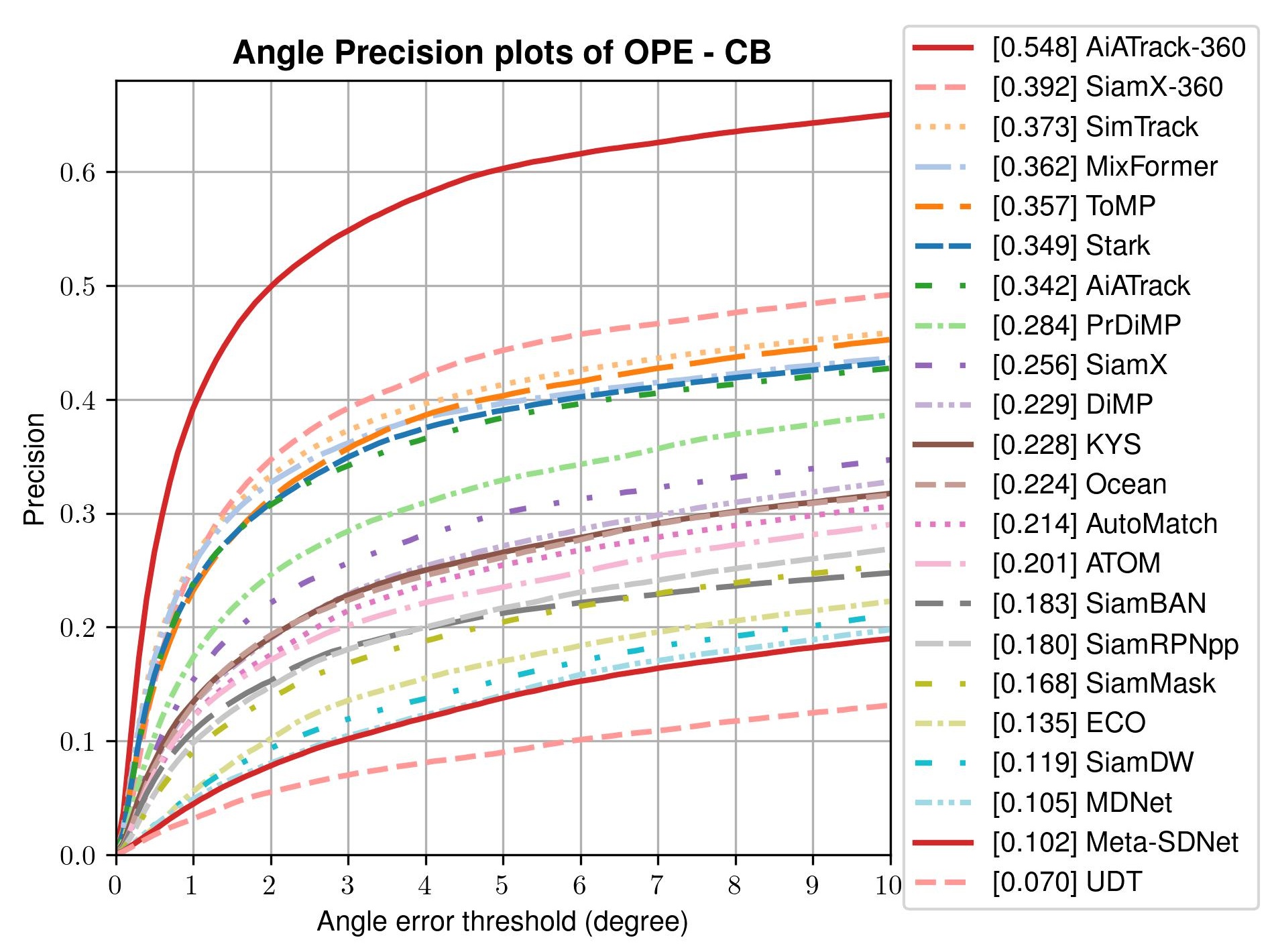}}
    \subfloat[\edited{(b) Fast Motion on the Sphere}]{\includegraphics[width=\imw\linewidth]{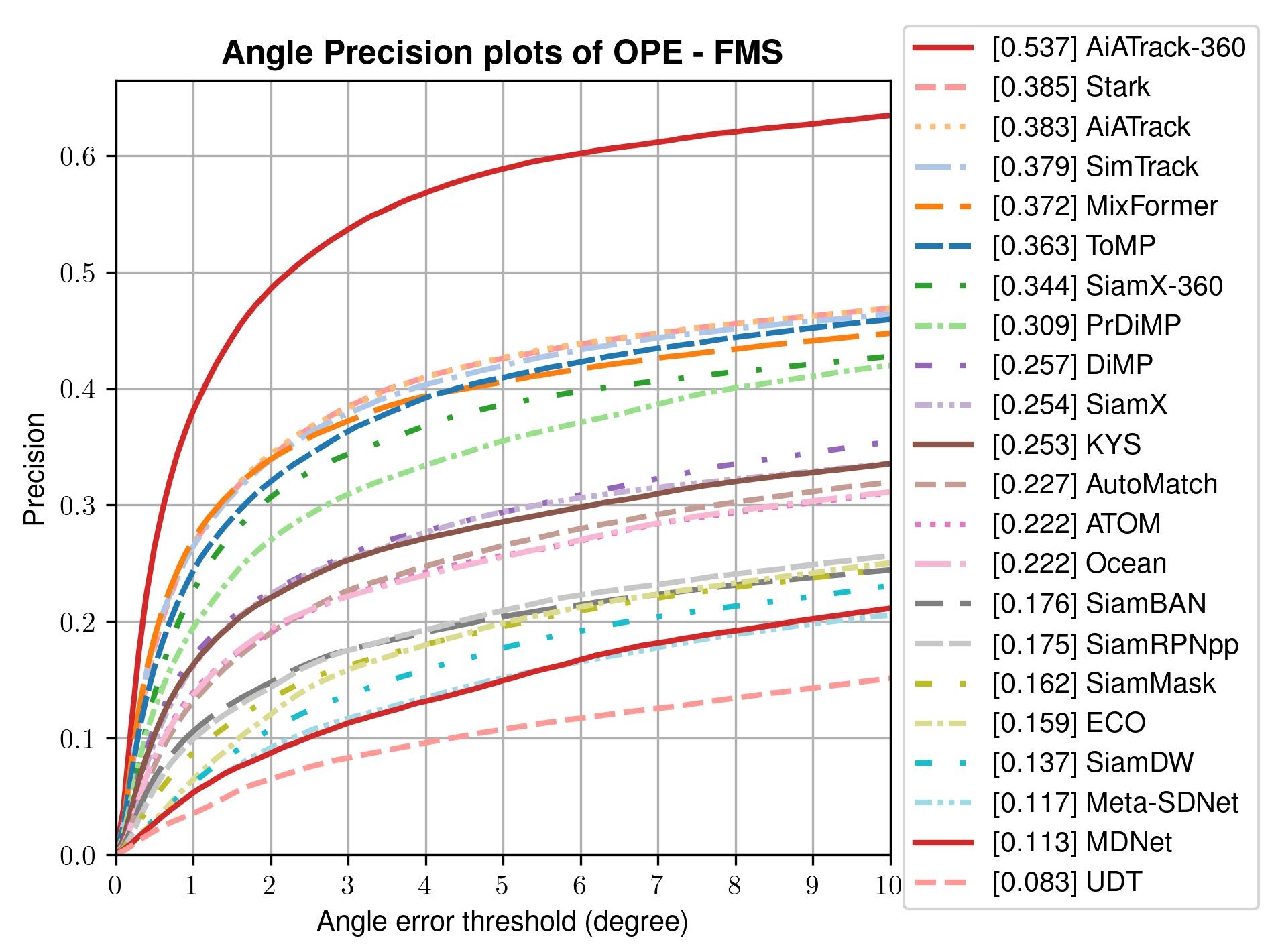}}
    \subfloat[\edited{(c) Latitude Variation}]{\includegraphics[width=\imw\linewidth]{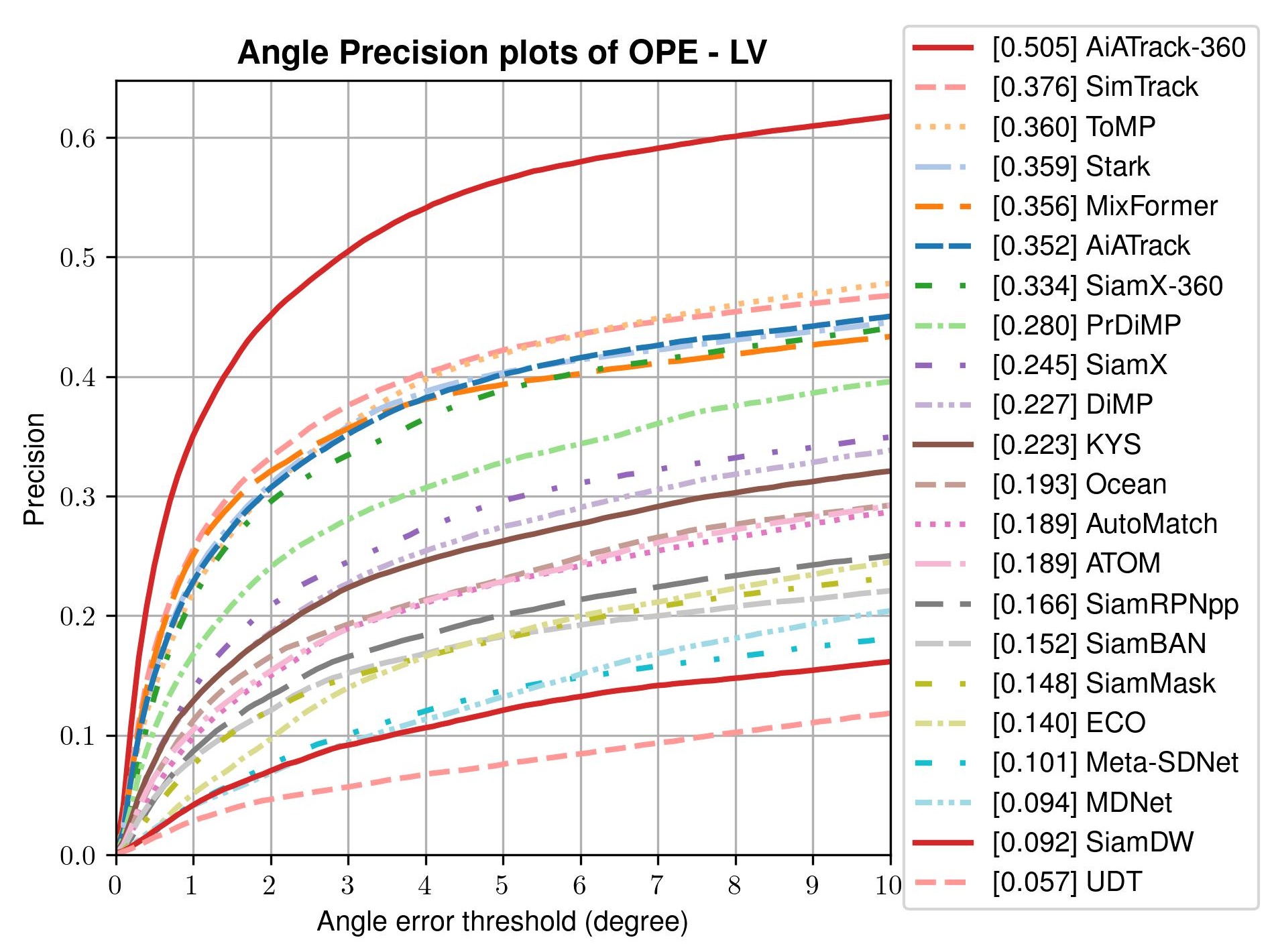}}
    \vspace{-0.2cm}
    \caption{Comparing BBox tracking performances of different trackers in terms of dual success rate and angle precision rate under the three distinct attributes of 360VOT.} 
    \label{fig:att_comp}
\end{figure*}

\noindent\textbf{Attribute-based performance}. Furthermore, we evaluate all trackers under 20 attributes in order to analyze different challenges faced by existing trackers. In Figure~\ref{fig:att_comp}, we plot the results on the videos with cross border (CB), fast motion on the sphere (FMS), and latitude variation (LV) attributes. These are the three exclusive and most challenging attributes of 360VOT. For complete results with other attributes, please refer to the supplementary. Compared to the overall performance, all trackers suffer performance degradation, especially on the CB and FMS attributes. For example, P$_{dual}$ of SimTrack decreases by 4.2\% and 5.3\% on CB and FMS respectively. However, the performance of AiATrack-360 still dominates on all three adverse attributes, while SiamX-360 also obtains stable performance gains.

\begin{table}[t]
    \centering
    \footnotesize
    \setlength{\tabcolsep}{4pt} 
    \begin{tabular}{lcccc }
    \toprule
    \multirow{2}{*}{Tracker}& \multicolumn{4}{c}{360VOT rBBox}\\
    \cmidrule(lr){2-5}
    &$\mbox{S}_{dual}$\footnotesize{(AUC)} &$\mbox{P}_{dual}$ &$\overline{\mbox{P}}_{dual}$\footnotesize{(AUC)}&$\mbox{P}_{angle}$\\
    \midrule
    SiamX-360     &0.205 &0.278 &0.278 &0.327\\
    AiATrack-360  &0.362 &0.449 &0.516 &0.535 \\
    \midrule
     &\multicolumn{2}{c}{360VOT BFoV}& \multicolumn{2}{c}{360VOT rBFoV} \\
     \cmidrule(lr){2-3}\cmidrule(lr){4-5}
     &$\mbox{S}_{sphere}$\footnotesize{(AUC)} &$\mbox{P}_{angle}$&$\mbox{S}_{sphere}$\footnotesize{(AUC)}&$\mbox{P}_{angle}$\\
     \midrule
    SiamX-360  &0.262 &0.327 &0.243 &0.323\\
    AiATrack-360  &0.548 &0.564 &0.426 &0.530 \\
    \bottomrule
    \end{tabular}    
    \vspace{-0.4em}
    \caption{Tracking performance based on other annotations of 360VOT using 360 tracking framework. }
    \label{tab:result2}
    \vskip -1em
\end{table}

\begin{table*}
    \centering\small
    \setlength{\tabcolsep}{1pt} 
    \def\imgw{0.19}
    \begin{tabular}{cccccc}
    {\rotatebox{90}{\parbox{0.07\linewidth}{\centering BBox}}}& \includegraphics[width=\imgw\linewidth]{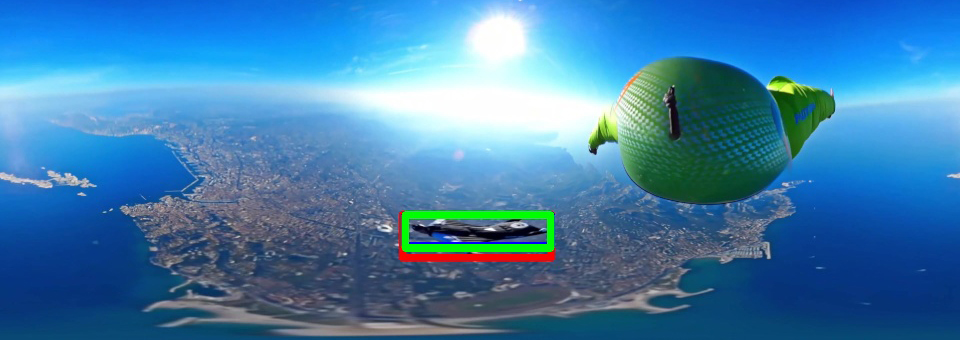}&
    \includegraphics[width=\imgw\linewidth]{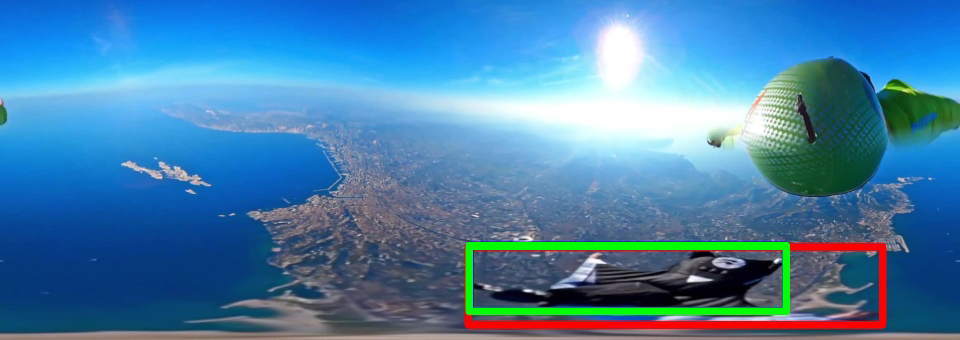}&
    \includegraphics[width=\imgw\linewidth]{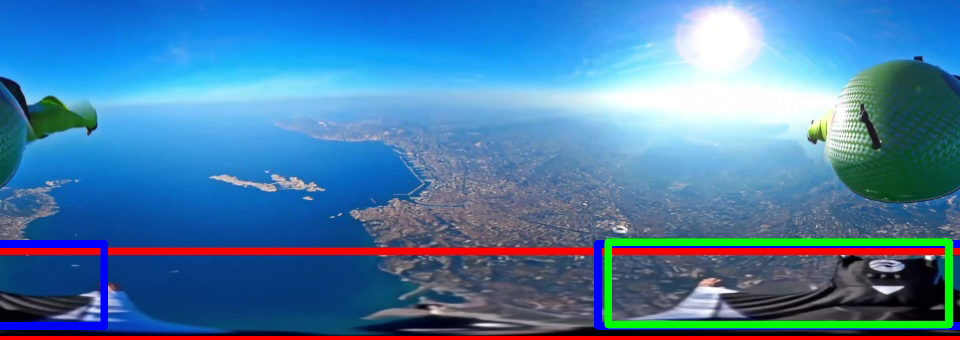}&
    \includegraphics[width=\imgw\linewidth]{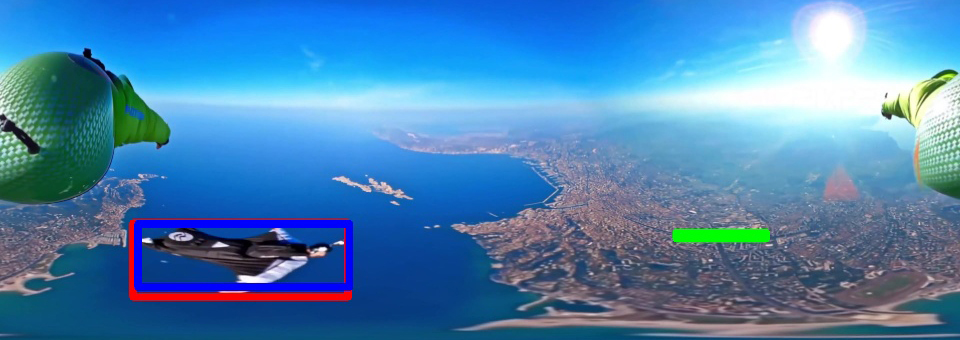}&
    \includegraphics[width=\imgw\linewidth]{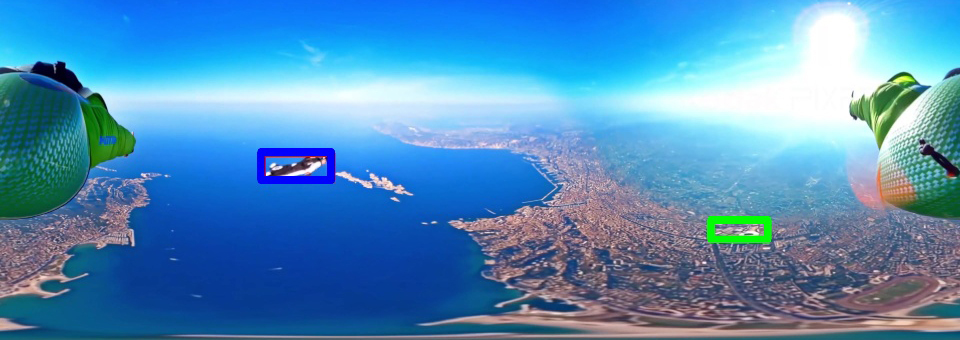}\\
    
    {\rotatebox{90}{\parbox{0.07\linewidth}{\centering rBBox}}}& \includegraphics[width=\imgw\linewidth]{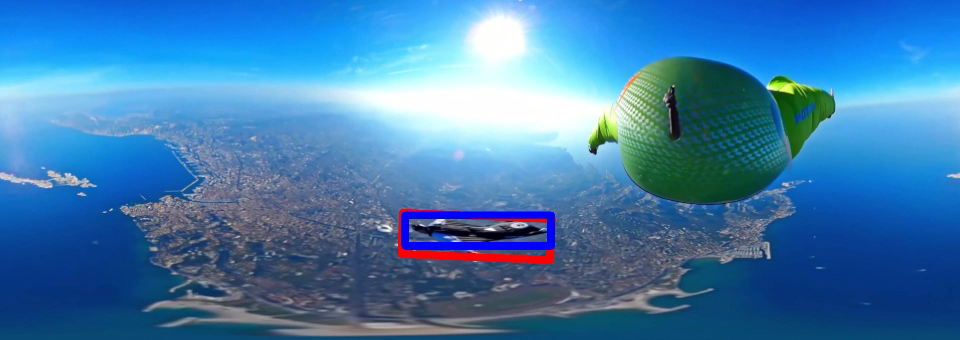}&
    \includegraphics[width=\imgw\linewidth]{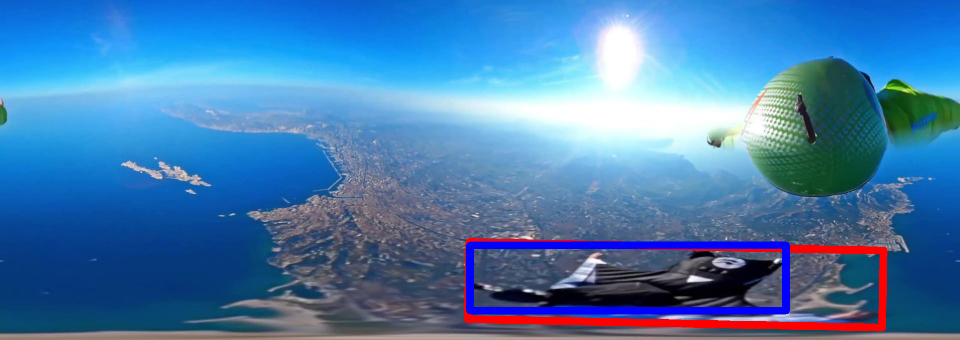}&
    \includegraphics[width=\imgw\linewidth]{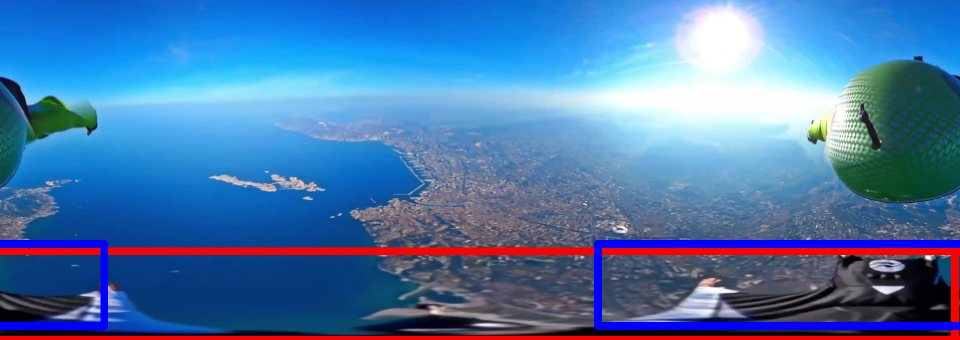}&
    \includegraphics[width=\imgw\linewidth]{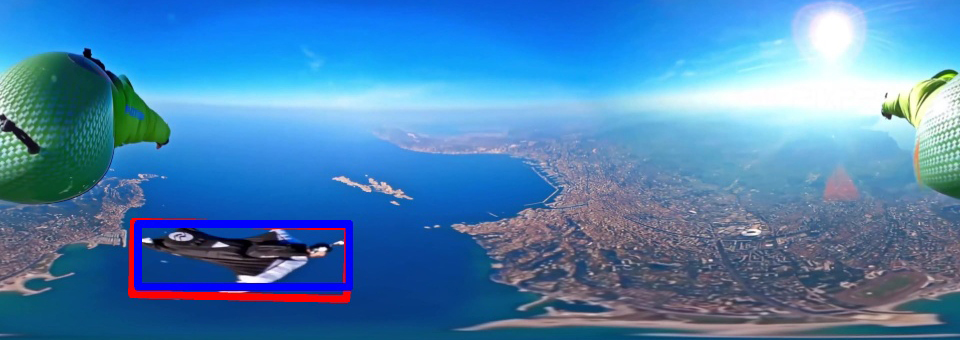}&
    \includegraphics[width=\imgw\linewidth]{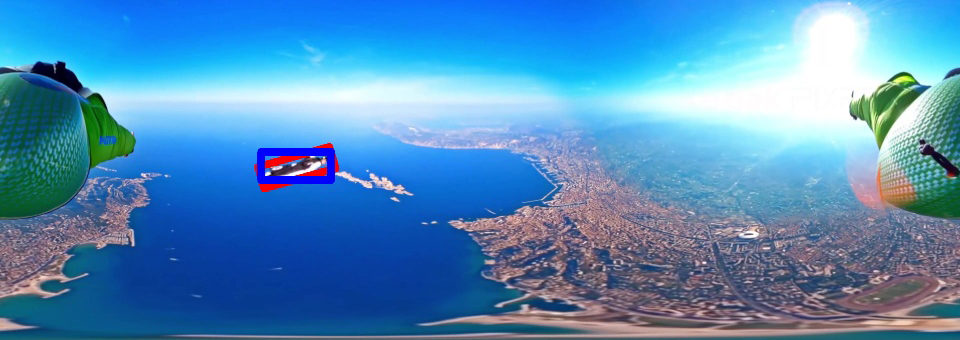} \\
    
    {\rotatebox{90}{\parbox{0.07\linewidth}{\centering BFoV}}}&\includegraphics[width=\imgw\linewidth]{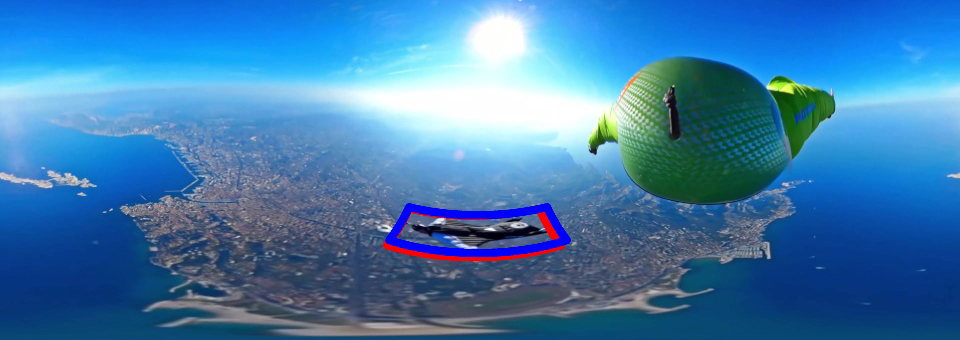}&
    \includegraphics[width=\imgw\linewidth]{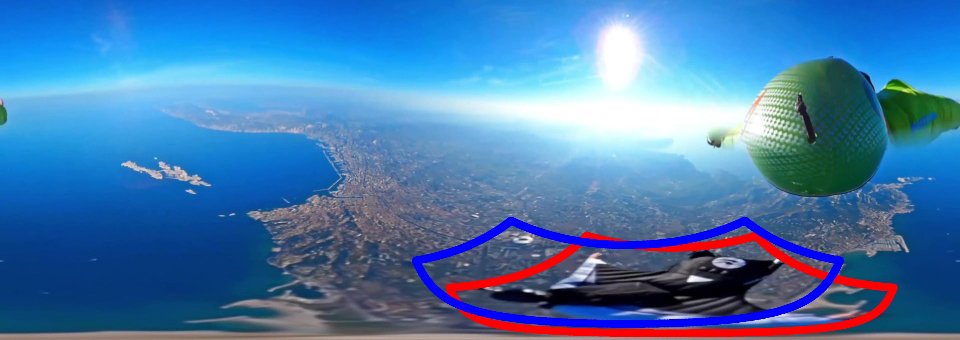}&
    \includegraphics[width=\imgw\linewidth]{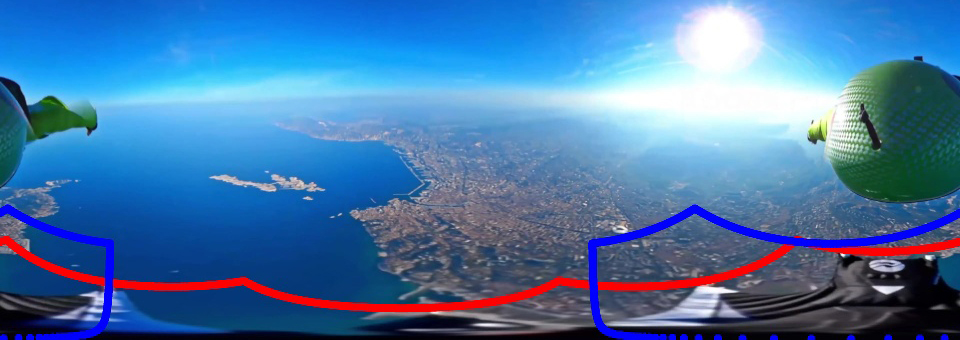}&
    \includegraphics[width=\imgw\linewidth]{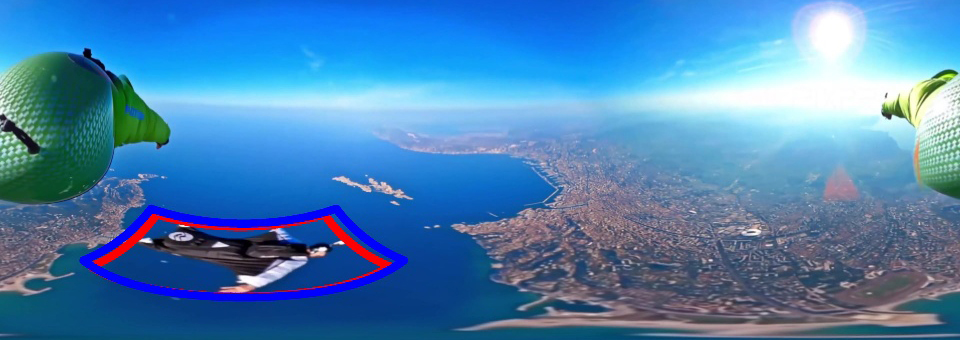}&
    \includegraphics[width=\imgw\linewidth]{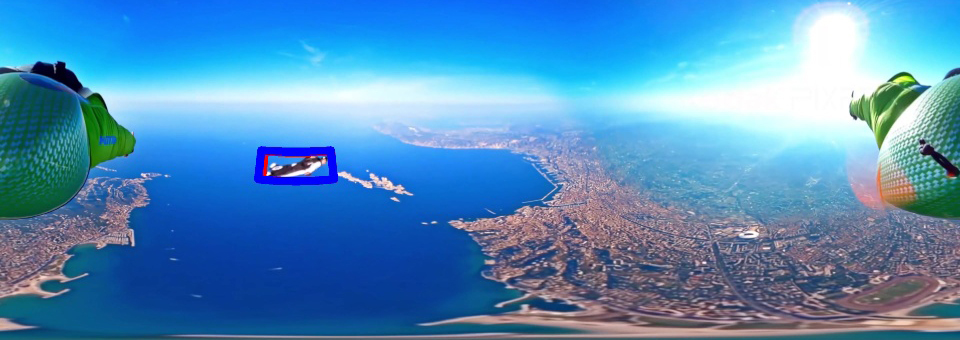}\\
    
    {\rotatebox{90}{\parbox{0.07\linewidth}{\centering rBFoV}}}&\includegraphics[width=\imgw\linewidth]{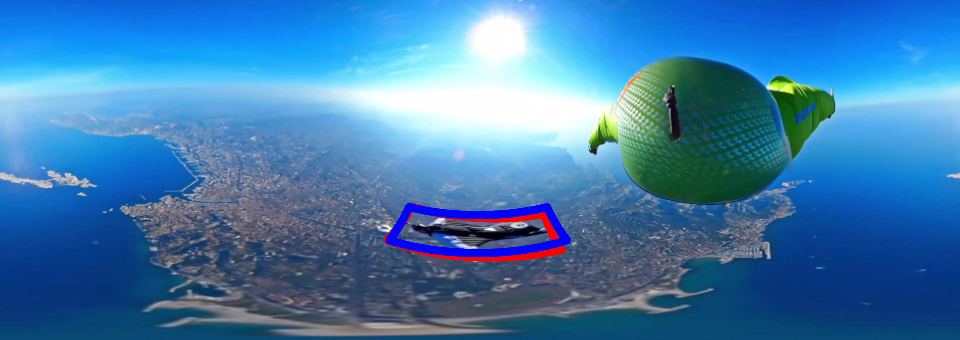}&
    \includegraphics[width=\imgw\linewidth]{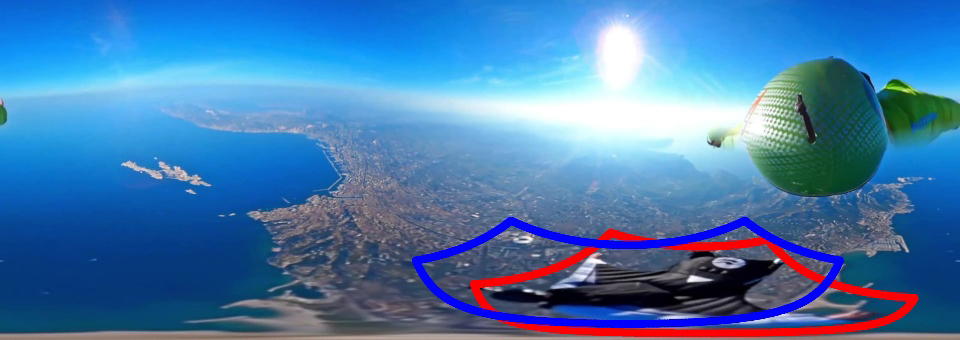}&
    \includegraphics[width=\imgw\linewidth]{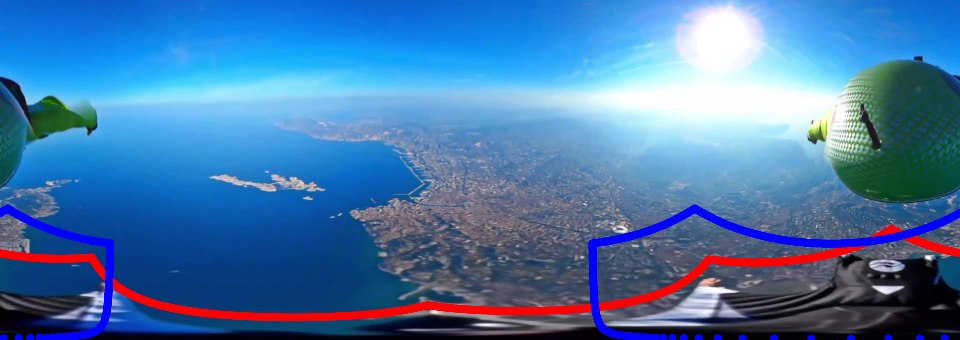}&
    \includegraphics[width=\imgw\linewidth]{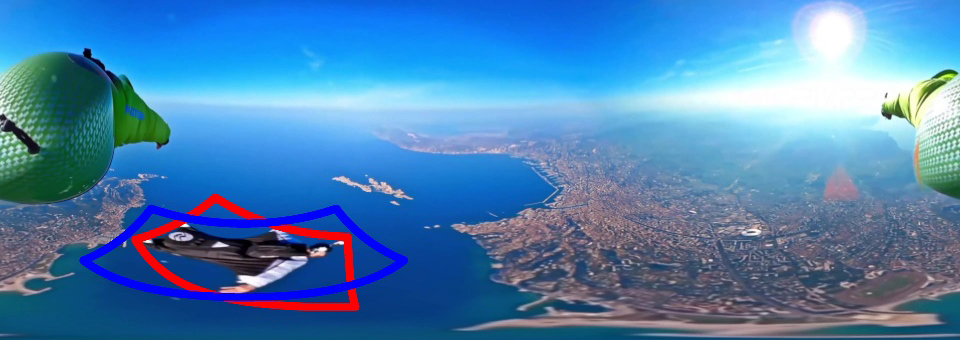}&
    \includegraphics[width=\imgw\linewidth]{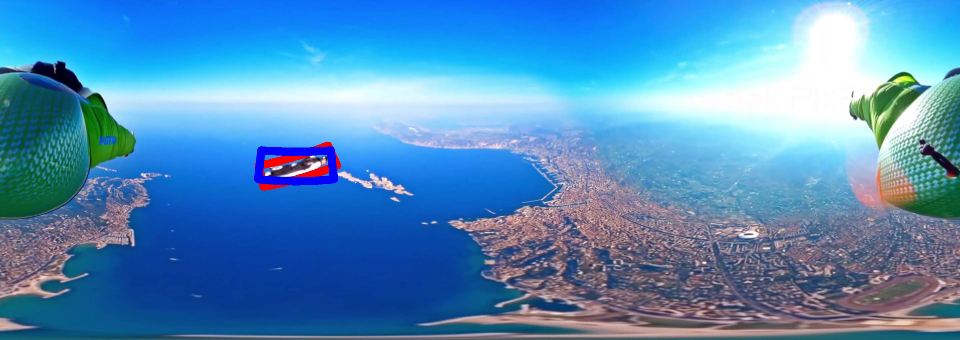}
    \end{tabular}
    \vspace{-0.2cm}
    \captionof{figure}{Qualitative results of the baseline on different representations. \textcolor{red}{Red} denotes the ground truth, and \textcolor{blue}{blue} denotes the results of AiATrack-360. The \textcolor{green}{green} in the first row denotes the results of AiATrack.}
    \label{fig:qual_result}
\end{table*}

\subsection{Performance based on other annotations}
Apart from BBox ground truth, we provide additional ground truth, including rBBox, BFoV, and rBFoV. As our 360 tracking framework can estimate approximate rBBox, BFoV, and rBFoV from local BBox predictions, we additionally evaluate performances of SiamX-360 and AiATrack-360 on these three representations (Table~\ref{tab:result2}). Compared with the results on BBox (Table~\ref{tab:result}), the performance on rBBox declines vastly. SiamX-360 and AiATrack-360 only achieve 0.205 and 0.362 S$_{dual}$ respectively. By contrast, the evaluation of BFoV and rBFoV has more reasonable and consistent numbers. 
In addition, we display visual results by AiATrack-360 and AiAtrack in Figure \ref{fig:qual_result}. 
AiATrack-360 can always follow and localize the target in challenging cases. Compared with (r)BBox, (r)BFoV can bind the target accurately with fewer irrelevant areas.
From the extensive evaluation, we can observe that using BFoV and rBFoV would be beneficial for object localization in omnidirectional scenes. 
While SiamX-360 and AiATrack-360 serve as new baselines to demonstrate this potential, developing new trackers which can directly predict rBBox, BFoV, and rBFoV will be an important future direction.


\section{Discussion and Conclusion}
The 360 tracking framework with existing tracker integration can, to some extent, succeed in omnidirectional visual object tracking, but there still remains much room for improvement. We want to discuss some promising directions here. 1) Data augmentation. The existing trackers are trained on the dataset of perspective images, while large-scale training data of 360$\degree$ images are lacking. During training, we can introduce projection distortion to augment the training data.   
2) Long-term omnidirectional tracking algorithms. The trackers enhanced by our tracking framework are technically still classified as short-term trackers .  As target occlusion is a noticeable attribute of 360VOT, the long-term tracker capable of target relocalization can perform \edited{better}. Nonetheless, how to effectively and efficiently search for targets on a whole 360$\degree$ image is a challenge.
3) New network architectures. SphereNet~\cite{spherenet} learns spherical representations for omnidirectional detection and classification, while DeepSphere~\cite{deepsphere} proposes a graph-based spherical CNN. The trackers exploiting these network architectures tailored for omnidirectional images may be able to extract better features and correlations for robust tracking.
By releasing 360VOT, we believe that the new dataset, representations, metrics, and benchmark can encourage more research and application of omnidirectional visual object tracking in both computer vision and robotics. 

\noindent\textbf{Acknowledgement.}
This research is partially supported by an internal grant from HKUST (R9429) and the Innovation and Technology Support Programme of the Innovation and Technology Fund (Ref: ITS/200/20FP).
We thank Xiaopeng Guo, Yipeng Zhu, Xiaoyu Mo, and Tsz-Chun Law for data collection and processing.

{\small
\bibliographystyle{ieee_fullname}
\bibliography{egbib}

\begin{thebibliography}{10}\itemsep=-1pt

\bibitem{dimp}
Goutam Bhat, Martin Danelljan, Luc~Van Gool, and Radu Timofte.
\newblock Learning discriminative model prediction for tracking.
\newblock In {\em Proceedings of the IEEE/CVF international conference on
  computer vision}, pages 6182--6191, 2019.

\bibitem{kys}
Goutam Bhat, Martin Danelljan, Luc Van~Gool, and Radu Timofte.
\newblock Know your surroundings: Exploiting scene information for object
  tracking.
\newblock In {\em Computer Vision--ECCV 2020: 16th European Conference,
  Glasgow, UK, August 23--28, 2020, Proceedings, Part XXIII 16}, pages
  205--221. Springer, 2020.

\bibitem{simtrack}
Boyu Chen, Peixia Li, Lei Bai, Lei Qiao, Qiuhong Shen, Bo Li, Weihao Gan, Wei
  Wu, and Wanli Ouyang.
\newblock Backbone is all your need: A simplified architecture for visual
  object tracking.
\newblock {\em arXiv preprint arXiv:2203.05328}, 2022.

\bibitem{siamban}
Zedu Chen, Bineng Zhong, Guorong Li, Shengping Zhang, and Rongrong Ji.
\newblock Siamese box adaptive network for visual tracking.
\newblock In {\em Proceedings of the IEEE/CVF Conference on Computer Vision and
  Pattern Recognition}, pages 6668--6677, 2020.

\bibitem{indoor360}
Shih-Han Chou, Cheng Sun, Wen-Yen Chang, Wan-Ting Hsu, Min Sun, and Jianlong
  Fu.
\newblock 360-indoor: Towards learning real-world objects in 360° indoor
  equirectangular images.
\newblock {\em 2020 IEEE Winter Conference on Applications of Computer Vision
  (WACV)}, 2020.

\bibitem{FlyingCars}
Benjamin Coors, Alexandru~Paul Condurache, and Andreas Geiger.
\newblock Spherenet: Learning spherical representations for detection and
  classification in omnidirectional images.
\newblock In {\em Proceedings of the European Conference on Computer Vision
  (ECCV)}, September 2018.

\bibitem{spherenet}
Benjamin Coors, Alexandru~Paul Condurache, and Andreas Geiger.
\newblock Spherenet: Learning spherical representations for detection and
  classification in omnidirectional images.
\newblock In {\em Proceedings of the European conference on computer vision
  (ECCV)}, pages 518--533, 2018.

\bibitem{mixformer}
Yutao Cui, Cheng Jiang, Limin Wang, and Gangshan Wu.
\newblock Mixformer: End-to-end tracking with iterative mixed attention.
\newblock In {\em Proceedings of the IEEE/CVF Conference on Computer Vision and
  Pattern Recognition}, pages 13608--13618, 2022.

\bibitem{uiou}
Feng Dai, Bin Chen, Hang Xu, Yike Ma, Xiaodong Li, Bailan Feng, Peng Yuan,
  Chenggang Yan, and Qiang Zhao.
\newblock Unbiased iou for spherical image object detection.
\newblock In {\em Proceedings of the AAAI Conference on Artificial
  Intelligence}, volume~36, pages 508--515, 2022.

\bibitem{atom}
Martin Danelljan, Goutam Bhat, Fahad~Shahbaz Khan, and Michael Felsberg.
\newblock Atom: Accurate tracking by overlap maximization.
\newblock In {\em Proceedings of the IEEE/CVF conference on computer vision and
  pattern recognition}, pages 4660--4669, 2019.

\bibitem{eco}
Martin Danelljan, Goutam Bhat, Fahad Shahbaz~Khan, and Michael Felsberg.
\newblock Eco: Efficient convolution operators for tracking.
\newblock In {\em Proceedings of the IEEE conference on computer vision and
  pattern recognition}, pages 6638--6646, 2017.

\bibitem{prdimp}
Martin Danelljan, Luc~Van Gool, and Radu Timofte.
\newblock Probabilistic regression for visual tracking.
\newblock In {\em Proceedings of the IEEE/CVF conference on computer vision and
  pattern recognition}, pages 7183--7192, 2020.

\bibitem{deepsphere}
Micha{\"e}l Defferrard, Martino Milani, Fr{\'e}d{\'e}rick Gusset, and
  Nathana{\"e}l Perraudin.
\newblock Deepsphere: a graph-based spherical cnn.
\newblock {\em arXiv preprint arXiv:2012.15000}, 2020.

\bibitem{uavdt}
Dawei Du, Yuankai Qi, Hongyang Yu, Yifan Yang, Kaiwen Duan, Guorong Li, Weigang
  Zhang, Qingming Huang, and Qi Tian.
\newblock The unmanned aerial vehicle benchmark: Object detection and tracking.
\newblock In {\em Proceedings of the European conference on computer vision
  (ECCV)}, pages 370--386, 2018.

\bibitem{trek150}
Matteo Dunnhofer, Antonino Furnari, Giovanni~Maria Farinella, and Christian
  Micheloni.
\newblock Is first person vision challenging for object tracking?
\newblock In {\em Proceedings of the IEEE/CVF International Conference on
  Computer Vision}, pages 2698--2710, 2021.

\bibitem{lasot}
Heng Fan, Liting Lin, Fan Yang, Peng Chu, Ge Deng, Sijia Yu, Hexin Bai, Yong
  Xu, Chunyuan Liao, and Haibin Ling.
\newblock Lasot: A high-quality benchmark for large-scale single object
  tracking.
\newblock In {\em Proceedings of the IEEE/CVF conference on computer vision and
  pattern recognition}, pages 5374--5383, 2019.

\bibitem{totb}
Heng Fan, Halady~Akhilesha Miththanthaya, Siranjiv~Ramana Rajan, Xiaoqiong Liu,
  Zhilin Zou, Yuewei Lin, Haibin Ling, et~al.
\newblock Transparent object tracking benchmark.
\newblock In {\em Proceedings of the IEEE/CVF International Conference on
  Computer Vision}, pages 10734--10743, 2021.

\bibitem{aiatrack}
Shenyuan Gao, Chunluan Zhou, Chao Ma, Xinggang Wang, and Junsong Yuan.
\newblock Aiatrack: Attention in attention for transformer visual tracking.
\newblock In {\em European Conference on Computer Vision}, pages 146--164.
  Springer, 2022.

\bibitem{kcf}
Jo{\~a}o~F Henriques, Rui Caseiro, Pedro Martins, and Jorge Batista.
\newblock High-speed tracking with kernelized correlation filters.
\newblock {\em IEEE transactions on pattern analysis and machine intelligence},
  37(3):583--596, 2014.

\bibitem{360vo}
Huajian Huang and Sai-Kit Yeung.
\newblock 360vo: Visual odometry using a single 360 camera.
\newblock In {\em 2022 International Conference on Robotics and Automation
  (ICRA)}, pages 5594--5600. IEEE, 2022.

\bibitem{siamx2022}
Huajian Huang and Sai-Kit Yeung.
\newblock Siamx: An efficient long-term tracker using cross-level feature
  correlation and adaptive tracking scheme.
\newblock In {\em International Conference on Robotics and Automation (ICRA)}.
  IEEE, 2022.

\bibitem{got10k}
Lianghua Huang, Xin Zhao, and Kaiqi Huang.
\newblock Got-10k: A large high-diversity benchmark for generic object tracking
  in the wild.
\newblock {\em IEEE transactions on pattern analysis and machine intelligence},
  43(5):1562--1577, 2019.

\bibitem{nfs}
Hamed Kiani~Galoogahi, Ashton Fagg, Chen Huang, Deva Ramanan, and Simon Lucey.
\newblock Need for speed: A benchmark for higher frame rate object tracking.
\newblock In {\em Proceedings of the IEEE International Conference on Computer
  Vision}, pages 1125--1134, 2017.

\bibitem{VOT}
Matej Kristan, Jiri Matas, Ale\v{s} Leonardis, Tomas Vojir, Roman Pflugfelder,
  Gustavo Fernandez, Georg Nebehay, Fatih Porikli, and Luka \v{C}ehovin.
\newblock A novel performance evaluation methodology for single-target
  trackers.
\newblock {\em IEEE Transactions on Pattern Analysis and Machine Intelligence},
  38(11):2137--2155, Nov 2016.

\bibitem{nus-pro}
Annan Li, Min Lin, Yi Wu, Ming-Hsuan Yang, and Shuicheng Yan.
\newblock Nus-pro: A new visual tracking challenge.
\newblock {\em IEEE transactions on pattern analysis and machine intelligence},
  38(2):335--349, 2015.

\bibitem{siamrpn++}
Bo Li, Wei Wu, Qiang Wang, Fangyi Zhang, Junliang Xing, and Junjie Yan.
\newblock Siamrpn++: Evolution of siamese visual tracking with very deep
  networks.
\newblock In {\em Proceedings of the IEEE/CVF conference on computer vision and
  pattern recognition}, pages 4282--4291, 2019.

\bibitem{dtb70}
Siyi Li and Dit-Yan Yeung.
\newblock Visual object tracking for unmanned aerial vehicles: A benchmark and
  new motion models.
\newblock In {\em Proceedings of the AAAI Conference on Artificial
  Intelligence}, volume~31, 2017.

\bibitem{tc128}
Pengpeng Liang, Erik Blasch, and Haibin Ling.
\newblock Encoding color information for visual tracking: Algorithms and
  benchmark.
\newblock {\em IEEE transactions on image processing}, 24(12):5630--5644, 2015.

\bibitem{tomp}
Christoph Mayer, Martin Danelljan, Goutam Bhat, Matthieu Paul, Danda~Pani
  Paudel, Fisher Yu, and Luc Van~Gool.
\newblock Transforming model prediction for tracking.
\newblock In {\em Proceedings of the IEEE/CVF conference on computer vision and
  pattern recognition}, pages 8731--8740, 2022.

\bibitem{uav123}
Matthias Mueller, Neil Smith, and Bernard Ghanem.
\newblock A benchmark and simulator for uav tracking.
\newblock In {\em Computer Vision--ECCV 2016: 14th European Conference,
  Amsterdam, The Netherlands, October 11--14, 2016, Proceedings, Part I 14},
  pages 445--461. Springer, 2016.

\bibitem{trackingnet}
Matthias Muller, Adel Bibi, Silvio Giancola, Salman Alsubaihi, and Bernard
  Ghanem.
\newblock Trackingnet: A large-scale dataset and benchmark for object tracking
  in the wild.
\newblock In {\em Proceedings of the European conference on computer vision
  (ECCV)}, pages 300--317, 2018.

\bibitem{mdnet}
Hyeonseob Nam and Bohyung Han.
\newblock Learning multi-domain convolutional neural networks for visual
  tracking.
\newblock In {\em The IEEE Conference on Computer Vision and Pattern
  Recognition (CVPR)}, June 2016.

\bibitem{meta}
Eunbyung Park and Alexander~C Berg.
\newblock Meta-tracker: Fast and robust online adaptation for visual object
  trackers.
\newblock In {\em Proceedings of the European Conference on Computer Vision
  (ECCV)}, pages 569--585, 2018.

\bibitem{YT-BB}
Esteban Real, Jonathon Shlens, Stefano Mazzocchi, Xin Pan, and Vincent
  Vanhoucke.
\newblock Youtube-boundingboxes: A large high-precision human-annotated data
  set for object detection in video.
\newblock In {\em proceedings of the IEEE Conference on Computer Vision and
  Pattern Recognition}, pages 5296--5305, 2017.

\bibitem{alov300}
Arnold~WM Smeulders, Dung~M Chu, Rita Cucchiara, Simone Calderara, Afshin
  Dehghan, and Mubarak Shah.
\newblock Visual tracking: An experimental survey.
\newblock {\em IEEE transactions on pattern analysis and machine intelligence},
  36(7):1442--1468, 2013.

\bibitem{erp}
John~P Snyder.
\newblock {\em Flattening the earth: two thousand years of map projections}.
\newblock University of Chicago Press, 1997.

\bibitem{ritm2022}
Konstantin Sofiiuk, Ilya~A Petrov, and Anton Konushin.
\newblock Reviving iterative training with mask guidance for interactive
  segmentation.
\newblock In {\em 2022 IEEE International Conference on Image Processing
  (ICIP)}, pages 3141--3145. IEEE, 2022.

\bibitem{hrnet2019}
Ke Sun, Bin Xiao, Dong Liu, and Jingdong Wang.
\newblock Deep high-resolution representation learning for human pose
  estimation.
\newblock In {\em CVPR}, 2019.

\bibitem{OxUvA}
Jack Valmadre, Luca Bertinetto, Joao~F Henriques, Ran Tao, Andrea Vedaldi,
  Arnold~WM Smeulders, Philip~HS Torr, and Efstratios Gavves.
\newblock Long-term tracking in the wild: A benchmark.
\newblock In {\em Proceedings of the European conference on computer vision
  (ECCV)}, pages 670--685, 2018.

\bibitem{hrnet2020}
Jingdong Wang, Ke Sun, Tianheng Cheng, Borui Jiang, Chaorui Deng, Yang Zhao,
  Dong Liu, Yadong Mu, Mingkui Tan, Xinggang Wang, et~al.
\newblock Deep high-resolution representation learning for visual recognition.
\newblock {\em IEEE transactions on pattern analysis and machine intelligence},
  43(10):3349--3364, 2020.

\bibitem{UDT}
Ning Wang, Yibing Song, Chao Ma, Wengang Zhou, Wei Liu, and Houqiang Li.
\newblock Unsupervised deep tracking.
\newblock In {\em The IEEE Conference on Computer Vision and Pattern
  Recognition (CVPR)}, 2019.

\bibitem{siammask}
Qiang Wang, Li Zhang, Luca Bertinetto, Weiming Hu, and Philip~HS Torr.
\newblock Fast online object tracking and segmentation: A unifying approach.
\newblock In {\em Proceedings of the IEEE/CVF conference on Computer Vision and
  Pattern Recognition}, pages 1328--1338, 2019.

\bibitem{otb50}
Yi Wu, Jongwoo Lim, and Ming-Hsuan Yang.
\newblock Online object tracking: A benchmark.
\newblock In {\em Proceedings of the IEEE conference on computer vision and
  pattern recognition}, pages 2411--2418, 2013.

\bibitem{otb100}
Yi Wu, Jongwoo Lim, and Ming-Hsuan Yang.
\newblock Object tracking benchmark.
\newblock {\em IEEE Transactions on Pattern Analysis and Machine Intelligence},
  37(9):1834--1848, 2015.

\bibitem{PANDORA}
Hang Xu, Qiang Zhao, Yike Ma, Xiaodong Li, Peng Yuan, Bailan Feng, Chenggang
  Yan, and Feng Dai.
\newblock Pandora: A panoramic detection dataset for object with orientation.
\newblock In {\em ECCV}, 2022.

\bibitem{stark}
Bin Yan, Houwen Peng, Jianlong Fu, Dong Wang, and Huchuan Lu.
\newblock Learning spatio-temporal transformer for visual tracking.
\newblock In {\em Proceedings of the IEEE/CVF international conference on
  computer vision}, pages 10448--10457, 2021.

\bibitem{osv2019}
Dawen Yu and Shunping Ji.
\newblock Grid based spherical cnn for object detection from panoramic images.
\newblock {\em Sensors}, 19(11):2622, 2019.

\bibitem{automatch}
Zhipeng Zhang, Yihao Liu, Xiao Wang, Bing Li, and Weiming Hu.
\newblock Learn to match: Automatic matching network design for visual
  tracking.
\newblock In {\em Proceedings of the IEEE/CVF International Conference on
  Computer Vision}, pages 13339--13348, 2021.

\bibitem{SiamDW}
Zhipeng Zhang and Houwen Peng.
\newblock Deeper and wider siamese networks for real-time visual tracking.
\newblock In {\em The IEEE Conference on Computer Vision and Pattern
  Recognition (CVPR)}, June 2019.

\bibitem{ocean}
Zhipeng Zhang, Houwen Peng, Jianlong Fu, Bing Li, and Weiming Hu.
\newblock Ocean: Object-aware anchor-free tracking.
\newblock In {\em Computer Vision--ECCV 2020: 16th European Conference,
  Glasgow, UK, August 23--28, 2020, Proceedings, Part XXI 16}, pages 771--787.
  Springer, 2020.

\bibitem{spherical_criteria}
Pengyu Zhao, Ansheng You, Yuanxing Zhang, Jiaying Liu, Kaigui Bian, and Yunhai
  Tong.
\newblock Spherical criteria for fast and accurate 360° object detection.
\newblock {\em Proceedings of the AAAI Conference on Artificial Intelligence},
  34(07):12959–12966, 2020.

\bibitem{desiamrpn}
Zheng Zhu, Qiang Wang, Bo Li, Wei Wu, Junjie Yan, and Weiming Hu.
\newblock Distractor-aware siamese networks for visual object tracking.
\newblock In {\em Proceedings of the European conference on computer vision
  (ECCV)}, pages 101--117, 2018.

\end{thebibliography}
}

\clearpage
\appendix

In this supplementary, we first demonstrate the proposed 360 tracking framework in detail. Then in Sec.~\ref{sec:supp_collect}, we detail the data collection criteria and categorization. In Sec.~\ref{sec:supp_anno}, we provide more information on annotation including the segmentation toolkit and conversion algorithms from masks to bounding boxes. Finally, we show a performance comparison between tangent BFoV and our extended BFoV.

\begin{table*}
    \centering
    \footnotesize
	\begin{tabular}{r||cccccccccccc} 
    \hline
	Benchmark & Videos & \makecell{Total\\frames} &\makecell{Min\\frames} & \makecell{Mean\\frames} & \makecell{Median\\frames} & \makecell{Max\\frames} & \makecell{Object\\classes} &Attr. &Annotation &Feature &Year\\
    \hline \hline
    {ALOV300}\cite{alov300}&314&152K&19&483&276&5,975&64&14&{sparse BBox}&diverse scenes&2013\\
    {OTB100}\cite{otb100}&100&81K&71&590&393&3,872&16&11&{dense BBox}&short-term&2015\\
    {NUS-PRO}\cite{nus-pro}&365&135K&146&371&300&5,040&8&12&{dense BBox}&occlusion-level&2015\\
    {TC128}\cite{tc128}&129&55K&71&429&365&3,872&27&11&{dense BBox}&{color enhanced}&2015\\
    {UAV123}\cite{uav123}&123&113K&109&915&882&3,085&9&12&{dense BBox}&UAV&2016\\
    {DTB70}\cite{dtb70}&70&16K&68&225&202&699&29&11&{dense BBox}&UAV&2016\\
    {NfS}\cite{nfs}&100&383K&169&3,830&2,448&20,665&17&9&{dense BBox}&high FPS&2017\\
    {UAVDT}\cite{uavdt}&100&78K&82&778&602&2,969&27&14&{sparse BBox}&UAV&2017\\
    
    {TrackingNet}$^*$\cite{trackingnet}&511&226K&96&441&390&2,368&27&15&{sparse BBox}&large scale&2018\\
    {OxUvA}\cite{OxUvA}&337&1.55M&900&4,260&2,628&37,440&22&6&{sparse BBox}&long-term&2018\\
    
    {LaSOT}$^*$\cite{lasot}&280&685K&1,000&2,448&2,102&9,999&85&14&{dense BBox}&{category balance}&2018\\
    {GOT-10k}$^*$\cite{got10k}&420&56K&29&127&100&920&84&6&{dense BBox}&generic&2019\\
    {TOTB}\cite{totb}&225&86K&126&381&389&500&15&12&{dense BBox} &transparent &2021\\
    {TREK-150}\cite{trek150}&150&97K&161&649&484&4,640&34&17&{dense BBox}&FPV&2021\\
    {VOT}\cite{VOT}&62&20K&41&321&242&1,500&37&9&{dense BBox}&annual&2022\\
    \hline 
    {360VOT}&120&113K&251&940&775&2,400&32&20&\makecell{dense (r)BBox\\\& (r)BFoV}&360$\degree$ images &2023\\
    \hline
    \end{tabular}
    \caption{Comparison of current popular benchmarks for visual single object tracking in the literature. $^*$ indicates that only the test set of each dataset is reported. 
    }
    \label{tab:benchmark}
\end{table*}

\begin{figure*}[t]
    \centering
    \includegraphics[width=1\linewidth]{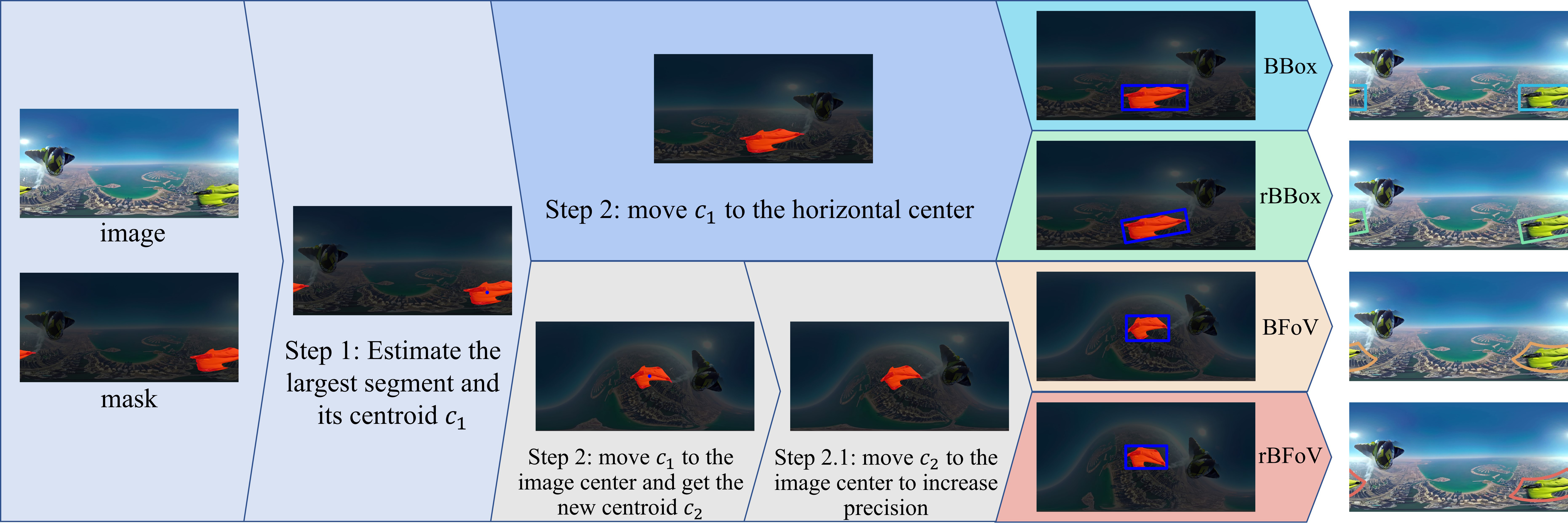}
    \caption{The 4 different annotations are generated by minimizing the bounding region of the object according to the segmentation.}
    \label{fig:mask2anno}
\end{figure*} 

\section{360 Tracking Framework}
We use a spherical camera model to depict the relationship between the 3D camera space $[X, Y, Z]$ and the 2D image space $[u, v]$. The projection function $\mathcal{F}$ is formulated as:
\begin{align}
    u &= (\frac{lon}{2\pi} + 0.5) * W = arctan(X/Z) ,\\
    v &= (-\frac{lat}{\pi}+0.5) * H = arctan(\frac{-Y}{\sqrt{X^2+Z^2}}) ,
\end{align}
where $-\pi<lon<\pi$ and $-\pi/2<lat<\pi/2$ denote the longitude and latitude in the spherical coordinate system respectively. W and H are the width and height of the 360$\degree$ image. As we mention in the main paper, a (r)BFoV is denoted as $[clon, clat, \theta, \phi, \gamma]$, where $clon$ and $clat$ represent the object center in the spherical coordinate system, $\theta$ and $\phi$ are the maximum bounding FoVs of the object, the rotation $\gamma$ of BFoV is zero. If we use a tangent plane $T\in\mathbb{R}^3$ to model the represented region of (r)BFoV, the corresponding region on 360$\degree$ is formulated as:
\begin{equation}
     I(\mbox{\footnotesize(r)BFoV} \,|\, \Omega) = \mathcal{F}(\mathcal{R}_y(clon)\cdot\mathcal{R}_x(clat)\cdot\mathcal{R}_z(\gamma)\cdot \Omega) .
\end{equation}
where,
\begin{align}
    \mathcal{R}_y(clon)&=\begin{bmatrix} cos(clon) & 0 &sin(clon)\\ 0& 1&0 \\-sin(clon) &0& cos(clon) \end{bmatrix}, \label{equ:roty}\\
    \mathcal{R}_x(clat)&=\begin{bmatrix} 1 & 0 &0\\ 0& cos(clat)&-sin(clat) \\0 &sin(clat)& cos(clat) \end{bmatrix},\label{equ:rotx}\\
    \mathcal{R}_z(\gamma)&= \begin{bmatrix} cos\gamma & -sin\gamma &0\\ sin\gamma& cos\gamma&0 \\0 &0& 1 \end{bmatrix},\label{equ:rotz}\\
    \Omega = T &= \begin{bmatrix} \mathbf{X}\\\mathbf{Y}\\\mathbf{Z}\end{bmatrix} =\begin{bmatrix}-tan(\theta/2):tan(\theta/2)\\-tan(\phi/2):tan(\phi/2)\\1\end{bmatrix},
\end{align}
To handle a large FoV, we extend the represented region of BFoV. When the FoV is larger than the threshold, e.g., 90$\degree$, the bounding region of BFoV becomes a surface segment $S\in\mathbb{R}^3$ of the unit sphere:
\begin{equation}
    S=\begin{bmatrix}cos(\Phi)sin(\Theta)\\-sin(\Phi)\\cos(\Phi)cos(\Theta)\end{bmatrix},\\
\end{equation}
where, $ \Phi \in [-\phi/2, \phi/2], \Theta\in [-\theta/2, \theta/2]$.
Therefore, the corresponding region of extended (r)BFoV on 360$\degree$ is formulated as:
\begin{equation}
    I(\mbox{\footnotesize(r)BFoV} \,|\, \Omega), \quad  \Omega = \begin{cases}
    T(\theta, \phi), &\theta<90\degree, \phi<90\degree\\
    S(\theta, \phi), &otherwise
    \end{cases} .
\end{equation}
Based on the $I$ which actually records the corresponding pixel coordinates of 360$\degree$, we can remap the 360$\degree$ image and extract a local search region to perform tracking which generates a BBox or rBBox prediction relative to the local region. After that, we still take advantage of $I$, converting the local prediction to obtain a global bounding region. To get the final (r)BBox prediction, we can calculate the minimum area (rotated) rectangle on the 360$\degree$. In addition, we can re-project the coordinates of the bounding region on the 360$\degree$ image to the spherical coordinates system and calculate the maximum bounding FoV for (r)BFoV.

\begin{figure}
    \centering
    \includegraphics[width=\linewidth]{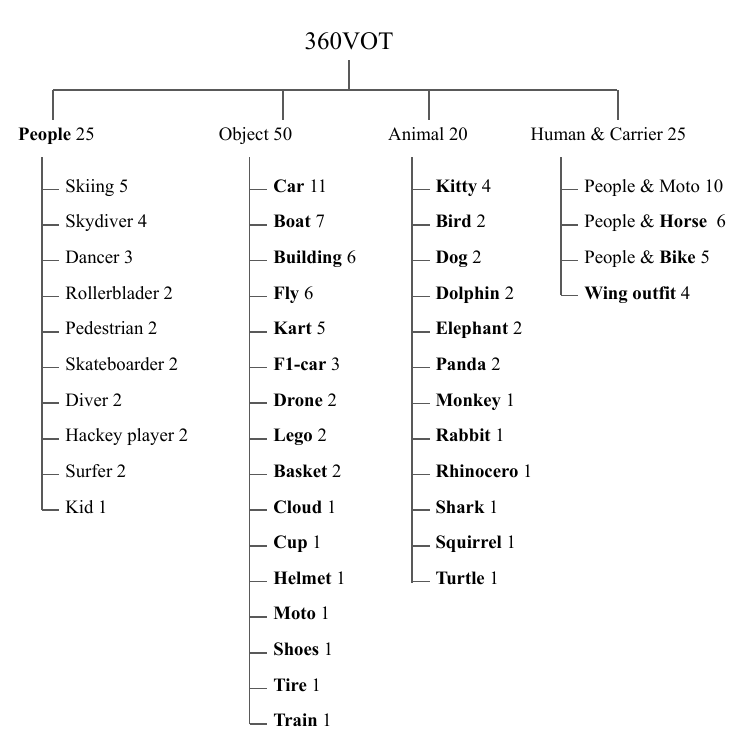}
    \caption{360VOT contains 120 sequences in diverse scenarios and 32 object categories which are denoted in bold. }
    \label{fig:360vot_categories}
\end{figure}
\section{Details of 360VOT Collection} \label{sec:supp_collect}
We manually collected videos from YouTube and self-recording captured some using a 360-degree camera. Four features were recorded for each sequence:
\textit{camera motion} (moving and stationary), \textit{target classes} (humans, animals, rigid objects and non-rigid objects), \textit{duration} (18 seconds to 75 minutes) and \textit{environment}. Specifically, the \textit{envonment} varies among indoor-outdoor, illumination (daylight, white light and night) and weather (cloudy, sunny and rainy). We ranked and filtered videos considering four criteria of tracking difficulty scale and some additional challenging cases. First, videos with the more adequate relative motion of the target and camera rank higher. Targets are preferably in a high degree of mobility, appearing in various locations across the frames rather than stationary. 
Second, videos with higher variability of the environment rank higher. The video background is supposed to be ever-changing across the video, such as with variations in lighting conditions. 
Third, videos with the target crossing frame boundaries rank higher. The object moving across frame boundaries is a distinct feature in panoramic videos. Finally, videos with a sufficient duration rank higher. A sufficient length of video provides a higher feasibility for the diversity of target movements and deformations, and possible disappearances, increasing tracking difficulties across the video. 

Eventually, the 360VOT benchmark dataset contains 120 sequences with about 113K frames in total. The minimum of frames is 251 while the average is 940. The types of targets can be placed in four major categories: \textit{People}, \textit{Object}, \textit{Animal} and \textit{Human \& Carrier}. In counting the class number of 360VOT, instead of subdividing the classes of humans, we describe it in a single broad category as \textit{People}. Since Horse and Bike classes in 360VOT always co-occur, we only classify them in \textit{Human \& Carrier} but not in \textit{Object} and \textit{Animal}. Finally, we consider it most appropriate to divide all targets into 32 categories, with the details of them are shown in Figure~\ref{fig:360vot_categories}. The comparison with current popular benchmarks is detailed in Table~\ref{tab:benchmark}

\section{Annotation} \label{sec:supp_anno}
Annotation is usually tedious and labor-intensive. For some existing benchmarks, they mentioned that they hired a large annotation team, more than 10 experts in the tracking domain, to manually label an enormous number of BBoxes over several months. However, such a strategy is not applicable for us to get 4 high-quality types of annotation. In addition, even though we can hire so many annotators with professional backgrounds, it is difficult to guarantee that the subjective annotations are optimal as the ground truth. The work~\cite{trek150} also reports the BBox annotation quality problem in the popular tracking benchmarks~\cite{otb100}. To obtain unbiased ground truth, we decide to segment the per-pixel target instance in each frame and then compute the corresponding (r)BBox and (r)BFoV from the resultant masks.

\subsection{Segmentation toolkit}S
\begin{figure}[t]
    \centering
    \includegraphics[width=1\linewidth]{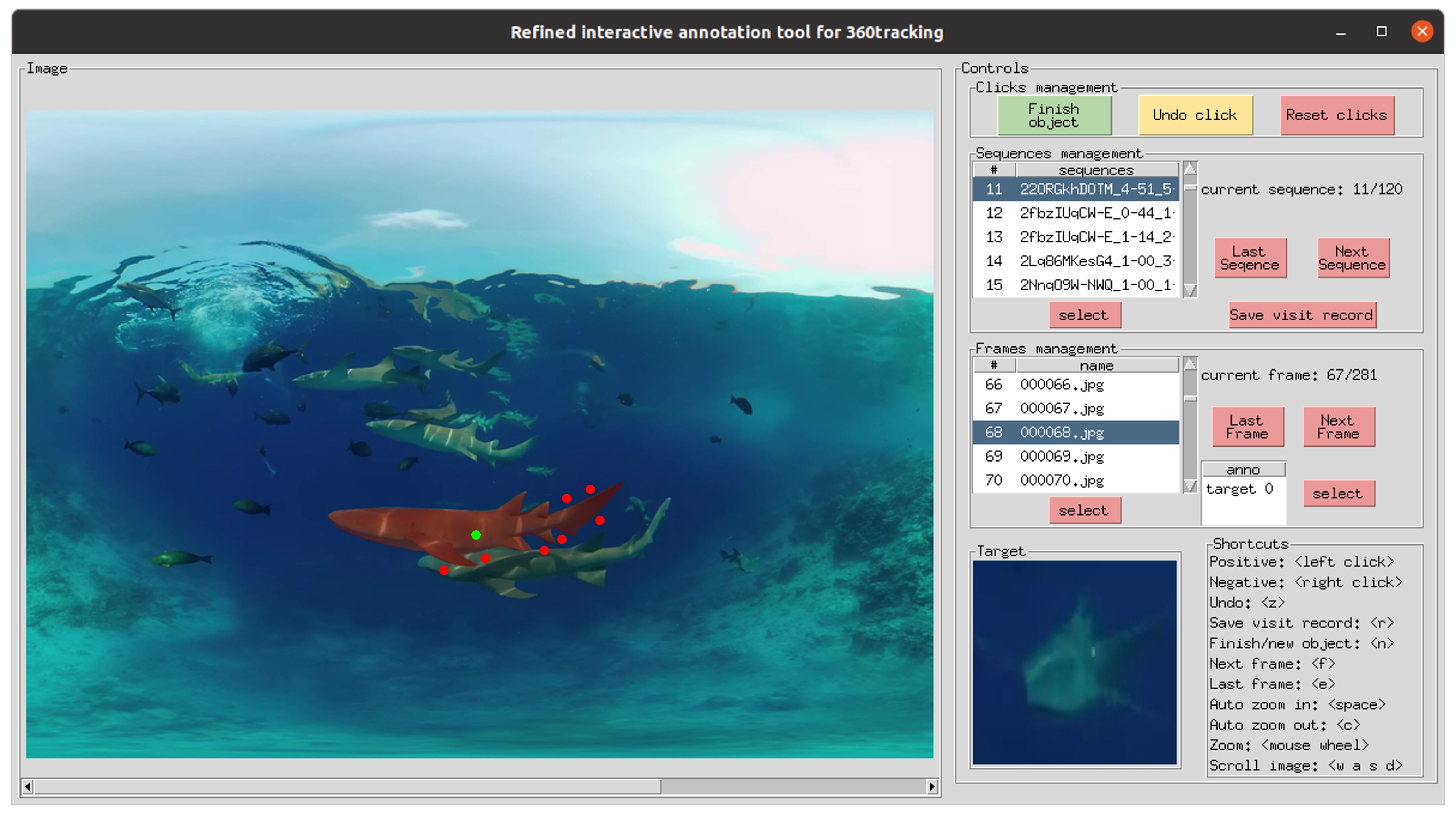}
    \caption{We take advantage of the click-based interactive segmentation model \cite{ritm2022} and develop a semi-automatic annotation tool to significantly increase the efficiency and attain high-quality annotation. 
    Annotators refine a segment via \textcolor{green}{green} positive and \textcolor{red}{red} negative clicks. }
    \label{fig:anno}
\end{figure}
To efficiently obtain fine-grained segmentation, we utilize a state-of-the-art tracker to get the initial positions of the targets with human online revision. The initial positions are then used by a semi-automatic segmentation toolkit to initialize the target object segmentation. The toolkit is based on a click-based interactive segmentation framework \cite{ritm2022}. The framework utilizes the HRNet-32~\cite{hrnet2019, hrnet2020} IT-M model trained on the COCO+LVIS dataset which can generate a complete segmentation on the instance with a few clicks. If the initial segmentation does not cover the target completely or contains elements not belonging to the target, positive (green) or negative (red) guiding points are manually added to generate a more accurate refined segmentation as shown in Figure~\ref{fig:anno}. 

\subsection{Mask to (r)BBox and (r)BFoV}
Essentially, the optimal annotation is to minimize the bounding area of the target. We can convert the mask to generate 4 types of unbiased ground truths. Specifically, since the masked target may span the left and right borders of the image, we first estimate the largest segment and then rotate the mask based on the centroid $c_1$ of the largest segment. To estimate BBox and rBBox, we only need to move the $c_1$ to the horizontal center of the mask image via Equ.~\ref{equ:roty} and then calculate the minimum area rectangle and rotate the rectangle respectively. However, for estimating the (r)BFoV, we need to rotate the mask to the mask center via Equ.~\ref{equ:roty} and ~\ref{equ:rotx} twice in order to reduce the distortion as much as possible. It is necessary to guarantee the accuracy of the estimations, especially for a large FoV. Next, we can calculate the bounding FoV to get the BFoV. But in terms of rBFoV estimation, we utilize the rotating calipers algorithm to estimate the rotation and then further rotate the mask via Equ.~\ref{equ:rotz} before calculating the bounding FoV. These processes are described in Algo.~\ref{alg:mask2bbox} and ~\ref{alg:mask2bfov}, and also illustrated in Figure~\ref{fig:mask2anno}.

\begin{algorithm}[t]
\footnotesize
\caption{Mask to (r)BBox}\label{alg:mask2bbox}
{\bf Input:} The mask $M$ and boolean value $needRotation$ 
\begin{algorithmic}
\State /*Step 1*/
\If{$M$ is empty} \State \Return $None$
\EndIf
\State $w_M \gets$ the width of the mask 
\State Convert Bound $M$ to set of polygons.
\State Estimate the largest segment and calculate the centroid $[x_1, y_1]$ in terms of the spherical coordinates, ${\theta_1, \phi_1}$ 
\State $\Delta x\gets x_1-w_M/2$
\State /*Step 2*/
\State Rotate $M$ by $\mathcal{R}_y(\theta_1)\sim$ Equ.~\ref{equ:roty}, giving $M_{R_1}$
\State /*Step 3*/
\If {$needRotation$}
    \State Bound $M_{R_1}$ by the minimum area rotated rectangle $[cx, cy, w, h, \gamma]$
\Else
    \State Bound $M_{R_1}$ by the minimum area rectangle $[cx, cy, w, h]$
    \State $\gamma\gets0$
\EndIf
\If{$w< w_M-1$}
    \State $cx\gets cx+\Delta x$
\Else
    \State $cx\gets w/2$
\EndIf
\State \Return $(cx,cy,w,h,\gamma)$
\end{algorithmic}
\end{algorithm}

\begin{algorithm}[t]
\footnotesize
\caption{Mask to (r)BFoV}\label{alg:mask2bfov}
{\bf Input:} The mask $M$ and boolean value $needRotation$ 
\begin{algorithmic}
\State /*Step 1*/
\If{$M$ is empty} 
    \State \Return $None$
\EndIf
\State Convert Bound $M$ to set of polygons.
\State Estimate the largest segment and calculate the centroid in terms of the spherical coordinates, ${\theta_1, \phi_1}$ 
\State /*Step 2*/
\State Rotate $M$ by $R_1=\mathcal{R}_y(\theta_1)\mathcal{R}_x(\phi_1)\sim$ Equ.~\ref{equ:roty} and~\ref{equ:rotx}, giving $M_{R_1}$
\State Calculate and convert the centroid of $M_{R_1}$ to the original $M$ in terms of the spherical coordinates, ${\theta_2, \phi_2}$ 
\State /*Step 2.1*/
\State Rotate $M$ by $R_2=\mathcal{R}_y(\theta_2)\mathcal{R}_x(\phi_2)$, giving $M_{R_2}$ 
\State Calculate the centroid $c_{R_2}$ , bounding width $w_{R_2}$, height $h_{R_2}$, and rotation $\gamma_{R_2}$ of $M_{R_2}$ by rotating calipers algorithm
\State Convert $c_{R_2}$ to the original $M$ and get the centroid in terms of the spherical coordinates, ${\theta_3, \phi_3}$ 
\If{$needRotation$}
    \If{$w_{R_2}>h_{R_2}$}
        \State $\gamma\gets\gamma_{R_2}$
    \Else
        \State $\gamma\gets\gamma_{R_2}-90$
    \EndIf
\Else
\State $\gamma\gets0$
\EndIf
\State /*Step 3*/
\State Rotate $M$ by $R_3=\mathcal{R}_y(\theta_3)\mathcal{R}_x(\phi_3)\mathcal{R}_z(\gamma)\sim$ Equ.~\ref{equ:roty}-~\ref{equ:rotz}, giving $M_{R_3}$

\State Calculate the range of longitude $[lon_{min}, lon_{max}]$ and latitude $[lat_{min}, lat_{max}]$ of $M_{R_3}$
\State Convert longitude center $(lon_{max}+lon_{min})/2$ of $M_{R_3}$  to orginal $M$, giving $clon$
\State Convert latitude center $(lat_{max}+lat_{min})/2$ of $M_{R_3}$  to orginal $M$, giving $clat$
\State $\theta\gets lon_{max}-lon_{min}$
\State $\phi\gets lat_{max}-lat_{min}$
\State \Return $(clon,clat,\theta,\phi,\gamma)$
\end{algorithmic}
\end{algorithm}

\section{More results}
\noindent\textbf{Tangent BFoV vs Extended BFoV}. As the FoV increases, the regions extracted by the tangent BFoV suffer extreme distortion, which would impact the tracking performance. To further verify the effectiveness of Extended BFoV, we conducted extra experiments, tracking based on the unwarped image of tangent BFoV. As reported in the main paper, the new baseline AiATrack-360 achieves 0.534 S$_{dual}$ on 360VOT BBox. However, if we conduct a search based on tangent BFoV, it encounters obvious degradation and only achieves 0.449 S$_{dual}$.

\end{document}